\documentclass[lettersize,journal]{IEEEtran}
\usepackage{amsmath,amsfonts}
\usepackage{algorithmic}
\usepackage{booktabs}
\usepackage{multirow}
\usepackage{array}
\usepackage{soul}
\usepackage{xcolor}
\usepackage{textcomp}
\usepackage{stfloats}
\usepackage{url}
\usepackage{verbatim}
\usepackage{graphicx}
\usepackage[caption=false,font=normalsize,labelfont=sf,textfont=sf]{subfig}
\usepackage{pgffor}

\def\BibTeX{{\rm B\kern-.05em{\sc i\kern-.025em b}\kern-.08em
    T\kern-.1667em\lower.7ex\hbox{E}\kern-.125emX}}
\usepackage{balance}
\usepackage[hidelinks]{hyperref}
\begin{document}
\title{DINO-CVA: A Multimodal Goal-Conditioned Vision-to-Action Model for Autonomous Catheter Navigation}

\author{Pedram Fekri, Majid Roshanfar, Samuel Barbeau, Seyedfarzad Famouri, Thomas Looi, \\Dale Podolsky, Mehrdad Zadeh, and Javad Dargahi

\thanks{P. Fekri, S. Famouri, and J. Dargahi are with the Gina Cody School of Engineering and Computer Science, Concordia University, Montreal, Canada.
{\tt\small p\_fekri@encs.concordia.ca; dargahi@encs.concordia.ca}}
\thanks{M. Roshanfar, T. Looi, and D. Podolsky are with The Wilfred and Joyce Posluns Centre for Image Guided Innovation \& Therapeutic Intervention (PCIGITI) at the Hospital for Sick Children (SickKids), Toronto, ON M5G 1E8, Canada.}
\thanks{S. Barbeau is with LIVIA, ÉTS Montréal, Canada, }
\thanks{M. Zadeh is with the Electrical and Computer Engineering Department, Kettering University, Flint, Michigan, USA.}
\thanks{This work has not been peer-reviewed. It is a preprint submitted to arXiv for early dissemination.\\
\copyright \ 2025 by the authors. This work is licensed under a non-exclusive license to arXiv.
}}


\maketitle

\begin{abstract}
Cardiac catheterization remains a cornerstone of minimally invasive interventions, yet it continues to rely heavily on manual operation.
Despite advances in robotic platforms, existing systems are predominantly follow–leader in nature, requiring continuous physician input and lacking intelligent autonomy.
This dependency contributes to operator fatigue, more radiation exposure, and variability in procedural outcomes.
This work moves towards autonomous catheter navigation by introducing DINO-CVA, a multimodal goal-conditioned behavior cloning framework. The proposed model fuses visual observations and joystick kinematics into a joint embedding space, enabling policies that are both vision-aware and kinematic-aware. Actions are predicted autoregressively from expert demonstrations, with goal conditioning guiding navigation toward specified destinations.
A robotic experimental setup with a synthetic vascular phantom was designed to collect multimodal datasets and evaluate performance. Results show that DINO-CVA achieves high accuracy in predicting actions, matching the performance of a kinematics-only baseline while additionally grounding predictions in the anatomical environment.
These findings establish the feasibility of multimodal, goal-conditioned architectures for catheter navigation, representing an important step toward reducing operator dependency and improving the reliability of catheter-based therapies.
\end{abstract}

\begin{IEEEkeywords}
Robotic catheterization, vision-to-action learning, autonomous navigation, teleoperation control, behavior cloning, imitation learning, goal-conditioned, DINO, multimodal transformer. 
\end{IEEEkeywords}

\section{Introduction}
\label{Introduction}
Minimally invasive surgery (MIS) uses small incisions or natural orifices to access internal organs, avoiding the large openings of traditional open surgery. By reducing the extent of skin and tissue disruption, MIS techniques significantly lower the physiological burden on patients. These procedures typically rely on advanced instruments and imaging technologies, such as endoscopes and catheters, which enable surgeons to operate with precision through narrow “keyhole” entry points~\cite{MIS-1}. Cardiac catheterization applies the principles of MIS by using a slender, flexible tube known as a catheter to traverse the vascular system and reach the heart. This approach enables both diagnostic and therapeutic procedures, such as angiography, electrophysiological mapping, and ablation, without the need for open-chest surgery. Modern catheters are equipped with steerable distal segments that are actuated by pull-wires or tendon-based mechanisms, which provide precise navigation through complex vascular pathways~\cite{keshavarz2025self}. This controllability allows operators to position the catheter tip accurately at targeted sites, including arrhythmogenic foci, where interventions such as stimulation or ablation can be carried out with high precision~\cite{catheterization-1}.

During cardiac catheterization, access is typically obtained through vascular entry sites such as the femoral or radial artery, and the catheter is advanced under continuous fluoroscopic monitoring. Fluoroscopy has become the gold standard for guiding these procedures, as it provides dynamic X-ray images that reveal the catheter’s position relative to surrounding anatomy. However, its routine use introduces important drawbacks. The dependence on ionizing radiation creates health risks for both patients and clinical staff, with cumulative exposure linked to malignancies, cataract formation, and radiation-induced skin damage. These risks are particularly concerning in complex ablation cases, where extended imaging times substantially increase total radiation dose.
Beyond radiation hazards, manual catheterization also places significant physical and cognitive demands on operators. Navigating the catheter through tortuous vasculature requires sustained attention, precise hand–eye coordination, and fine motor control, often maintained for hours at a time. These tasks are performed while wearing heavy lead aprons for radiation shielding, forcing clinicians to adopt static and ergonomically unfavorable positions. Over time, such working conditions contribute to operator fatigue and have been strongly associated with a high prevalence of musculoskeletal disorders among interventional cardiologists. Chronic back pain, neck strain, and joint injuries are frequently reported, underscoring the considerable occupational burden imposed by conventional catheterization practices~\cite{fatigue_1, fatigue_2, fatigue_3}.

Robotic-assisted catheterization platforms have been developed to address limitations of manual procedures such as radiation exposure, operator fatigue, and limited precision~\cite{roshanfar2025advanced}. Early systems like the Sensei Robotic System (Hansen Medical) and the Magellan Robotic System (Auris Health) enhanced catheter stability, reach, and steerability within tortuous vessels~\cite{RoboticsEndovascular, robotics2, robotics3}, while the CorPath GRX (Corindus/Siemens) enabled precise remote manipulation of guidewires, balloons, and stents during percutaneous interventions, reducing radiation exposure and demonstrating feasibility for tele-operated procedures~\cite{tele_corpath, corpath}. In parallel, magnetic navigation systems such as the Niobe (Stereotaxis) have leveraged externally generated magnetic fields to achieve superior catheter stability and mapping accuracy in electrophysiological ablation~\cite{magnet}, whereas simpler designs like the Amigo Remote Catheter System (Catheter Precision) provided safe and effective remote manipulation of conventional catheters with reduced operator radiation exposure~\cite{amigo}. Recent research prototypes are further advancing the field by incorporating force feedback and semi-autonomous control, including magnetorheological-fluid–based haptic feedback and platforms such as MAGiC RMN, which improve contact force regulation and demonstrate high success rates in arrhythmia ablation~\cite{magnofluid_1, magnofluid_2}.

Despite these technological advances, current robotic catheterization platforms remain fundamentally limited by their follow–leader architectures, which require constant physician control. Although these systems reduce radiation exposure and improve catheter stability, they do not alleviate the cognitive and physical workload of the operator, nor do they provide intelligent capabilities such as adaptive decision-making or autonomous navigation in complex anatomies. To address these limitations, there is a growing need for methods that can be seamlessly integrated into existing robotic platforms to enable semi-autonomous or fully autonomous catheter manipulation.

This work introduces an autonomous navigation system for cardiac catheterization, aimed towards reducing operator dependency and improving the reliability of catheter-based therapies. The proposed method is a multimodal vision–action model that learns from expert demonstrations in a learning-from-demonstration (LfD) framework by fusing visual observations with kinematic states. By moving beyond follow–leader configurations and incorporating intelligent navigation strategies, this approach represents an important step toward enabling autonomy that reflects expert practice. In the following section, related works on autonomous catheter navigation are reviewed, after which the details of the proposed method and its key contributions are presented.
\section{Related Work}
\label{related}
The pursuit of autonomy in robotic catheterization has recently gained momentum, with research efforts primarily centered around three cutting-edge learning paradigms: Reinforcement Learning (RL), Inverse Reinforcement Learning (IRL), and Imitation Learning (IL). The following sections present the state-of-the-art methodologies in each paradigm.

\subsection{Reinforcement Learning (RL)} 
RL has been widely explored for autonomous navigation and skill transfer in vascular and catheter-based interventions, leveraging trial-and-error interactions in simulated or physical environments to develop adaptive decision-making policies~\cite{RL-1, yolo5, RL-2, RL-3, RL-4, cathsim}. Studies have investigated diverse frameworks ranging from Deep Q-Network (DQN)-based navigation~\cite{RL-1} and Soft Actor-Critic (SAC) methods for real-time guidewire control~\cite{yolo5, RL-2} to Deep Deterministic Policy Gradient (DDPG)-driven virtual reality training platforms for skill transfer~\cite{RL-3}. More recent work integrates image guidance and advanced path planning~\cite{RL-4}, while open-source simulators such as CathSim facilitate reproducible RL training with algorithms like SAC and Proximal Policy Optimization (PPO)~\cite{cathsim}. 

\subsection{Imitation Learning (IL)} 
IL has attracted growing interest in robotic catheterization and minimally invasive interventions as it enables systems to acquire expert-level behaviors directly from demonstrations (Learning from Demonstration "LfD"), bypassing the need for handcrafted models or purely trial-and-error learning~\cite{IL-1, IL-2, IL-3, IL-4}. Early studies explored IL for tasks such as 3D/2D model-to-image registration in cardiac procedures, where neural networks learned to mimic expert alignment strategies to improve accuracy in image-guided interventions~\cite{IL-1}. Generative Adversarial Imitation Learning (GAIL) frameworks have further been used for autonomous catheter navigation and soft-tissue manipulation, often refined through Proximal Policy Optimization (PPO) to enhance precision, adaptability, and safety~\cite{IL-2, IL-3}. More recent work integrates IL with reinforcement learning, combining expert demonstrations with policy optimization and curriculum learning to improve robustness, collision avoidance, and generalization across complex anatomical environments~\cite{IL-4}. 

\subsection{Inverse Reinforcement Learning (IRL)} 
IRL extends beyond standard imitation learning by inferring the latent reward functions underlying expert demonstrations, enabling the development of adaptive and generalizable control policies rather than simply replicating observed behaviors~\cite{IRL-1, IRL-2}. In robotic catheterization, IRL has been applied to pre-operative path planning and autonomous navigation, where frameworks combining Behavioral Cloning (BC) and Generative Adversarial Imitation Learning (GAIL) inferred reward structures from expert trajectories to guide catheter movements in dynamic and patient-specific anatomies~\cite{IRL-1}. Other studies have used IRL to model expert decision-making in procedures such as mechanical thrombectomy, recovering reward functions from recorded state–action pairs and using them to train reinforcement learning policies via algorithms like Soft Actor-Critic (SAC), thereby enhancing precision, safety, and adaptability in complex surgical environments~\cite{IRL-2}.

\subsection{Challenges of Existing Paradigms}
Despite significant progress, existing learning paradigms face major challenges in robotic catheterization. RL suffers from sample inefficiency, sim-to-real gaps, reward engineering complexity, safety concerns, and instability in high-dimensional control settings. IL mitigates reward design issues but depends heavily on large expert datasets, struggles with covariate shift, lacks adaptability to novel anatomies, faces ambiguity in demonstrations, and does not explicitly optimize for safety or performance; moreover, most IL methods rely on kinematic or kinesthetic demonstrations with limited visual perception of the surgical scene or require external detectors such as YOLOv5 for scene understanding~\cite{farzad, farzad_2, fekri1, fekri2}. IRL attempts to infer latent rewards but encounters ambiguity in recovered objectives, computational burdens in high-dimensional spaces, and poor generalization to unseen clinical scenarios.

Addressing the challenges above, in this work, we propose a multimodal goal-conditioned behavior cloning model for autonomous catheter navigation. The model is inspired by DINO-WM in its architecture and is designed to learn from both visual features and the robot’s controller states (i.e., actions) over time \cite{dinowm}. Specifically, it functions as a multimodal vision–action model: given the video frames of the catheterization task together with the joystick states corresponding to the robot’s motors (translation, rotation, and knob), the model predicts the next action in an autoregressive manner. Unlike previous approaches that rely on object detection for catheter or vessel localization or on pre-operative trajectory planning, our ViT-based model learns the navigation policy by fusing visual cues and action states and processing them jointly over time with both temporal and spatial awareness. In contrast to DINO-WM, which leverages unsupervised strategies, our model is trained in a supervised manner using demonstration data collected from expert surgeons performing catheter insertion tasks \cite{dinowm}. To address the challenges of sample inefficiency and reward design inherent to reinforcement learning, our approach adopts a learning-from-demonstration (LfD) framework. However, unlike earlier LfD studies that primarily replicate kinematic or kinesthetic motions of experts, our model integrates visual understanding with kinematic features to capture a more human-like representation of expert decision-making. A goal-conditioning mechanism is incorporated to improve adaptability and generalization beyond conventional behavior cloning. Specifically, the model is conditioned on a frame representing the target location of the catheter within the vascular system, allowing it to plan trajectories by projecting learned behaviors toward the surgical goal.

We adopt IL rather than RL for cardiac catheterization due to the structural and procedural characteristics of the task. Unlike conventional RL settings, where exploration is essential for discovering optimal behaviors, the vascular anatomy in cardiac interventions is fixed and fully known prior to the procedure, thereby eliminating the need for extensive trial-and-error interaction with the environment. Moreover, expert demonstrations provide direct guidance on navigating through the vascular network toward clinically relevant targets, enabling the model to learn goal-directed control without the risk or inefficiency associated with exploration-based methods. By employing a goal-conditioned IL framework, the policy is explicitly conditioned on the desired anatomical endpoint, allowing it to accurately reproduce expert-like trajectories along the vascular pathways while leveraging the static nature of the environment to ensure both safety and efficiency.

Furthermore, because the proposed architecture captures expert demonstrations in both visual and kinematic spaces, it can be extended to IRL settings to assist with reward modeling. To support training and evaluation, we also designed and implemented a robotic platform equipped with a remote-control mechanism, enabling operators to steer a catheter within an artificial vascular system. This setup provided the necessary expert demonstration data for supervised training and empirical validation of the proposed model. To the best of our knowledge, this is the first framework to integrate goal-conditioned behavior cloning with visual–kinematic learning for autonomous catheter navigation, supported by a custom robotic platform for data collection and evaluation. For the sake of simplicity, we refer to the proposed framework as Dino-CVA (Dino Catheter Vision-Action model) throughout this work.
\section{Experimental Setup and Data Compilation}
\label{Experimental Setup and Data Compilation}
\label{setup}

Since commercial robotic catheterization platforms are not readily accessible for research purposes, we developed a simplified yet representative version to support demonstration, data compilation, and model validation. The system was designed to mimic the follow–leader paradigm of clinical robots, while remaining modular and cost-effective. A standard ablation catheter was integrated with a custom robotic actuation unit and a teleoperation interface, allowing the operator to control catheter motion remotely. Through an Xbox controller, user inputs were mapped directly to robotic actuators that translated, rotated, and steered the catheter tip. Real-time visual feedback was provided by a top-mounted camera, replicating the single-view setup commonly used in fluoroscopic navigation. In this way, the operator interacted with the platform by issuing joystick commands while relying solely on the camera feed for guidance, thereby closely reproducing the workflow of existing robotic systems within a controlled experimental environment. The robotic platform, interface software, and data compilation procedure are detailed in the following subsections.

\subsection{Robotic Platform for Catheter Navigation}
\label{Robotic Platform for Catheter Navigation}
\begin{figure*}
    \centering
    \includegraphics[width=\linewidth]{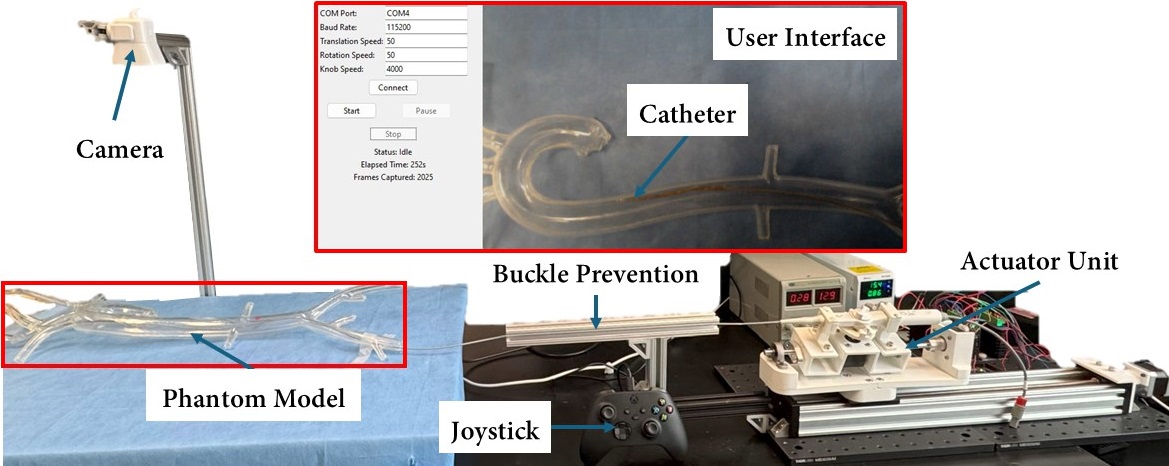}
    \caption{Overall experimental setup. The system includes a transparent vessel phantom placed under a top-mounted camera, a robotic platform with an actuator unit mounted on a linear rail, and a joystick controller for teleoperation. A buckle-prevention guide ensures stable catheter insertion. The graphical user interface allows parameter configuration and provides real-time visualization of the catheter within the phantom.}
    \label{fig:01}
\end{figure*}

\begin{figure}
    \centering
    \includegraphics[width=\linewidth]{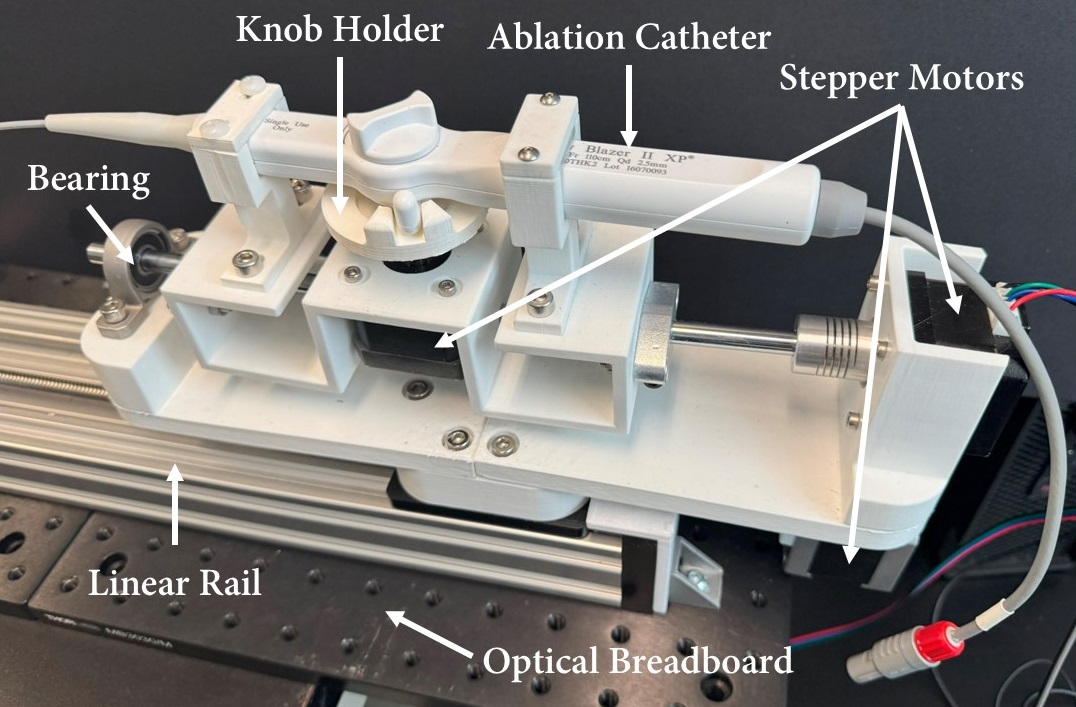}
    \caption{Close-up of the robotic platform. Three stepper motors, each driven by a microstepping driver, provide three DoFs: catheter translation, distal tip steering, and base rotation. The system is mounted on an optical breadboard, with 3D-printed fixtures securing the catheter during operation.}
    \label{fig:02}
\end{figure}

A transparent, commercially available synthetic vessel phantom was used to replicate the cardiac environment, enabling the insertion of an ablation catheter (Blazer XP, Boston Scientific, 110 cm length, OD 2.5 mm). The phantom was constructed at a 1:1 scale based on a medically validated human anatomical model. It was mounted on a tabletop, while a top-mounted camera provided a fixed overhead view of the procedure. This ensured that both the phantom and camera remained stable throughout catheterization. The robotic platform provided three degrees of freedom (DoFs), each actuated by a NEMA 17 stepper motor and driven by an independent TB6600 microstepping driver. The platform’s base was securely mounted on an optical breadboard to minimize vibration and prevent undesired motion. The three DoFs included: (1) a linear rail with 55~cm travel for catheter advancement and retraction, (2) a rotational stage directly coupled to the catheter knob for steering the distal tip, and (3) a base rotation module capable of rotating the entire catheter within $\pm 90^{\circ}$. Structural components were 3D printed to provide secure catheter fixation, modularity, and smooth motion during operation. A Logitech StreamCam (1080p, 60~fps) was mounted above the phantom to capture real-time navigation. To avoid catheter buckling during insertion, a hollow guiding rod was positioned at the center of the setup, through which the catheter was advanced. This configuration enabled precise, stable, and repeatable catheter control within the phantom. For teleoperation, an Xbox 360 wired controller was employed to command the three robotic DoFs. The overall setup is shown in Fig.~\ref{fig:01}, while a close-up view of the robotic platform is provided in Fig.~\ref{fig:02}.

\subsection{Interface Software}
\label{Interface Software}
A custom software environment was developed to enable real-time teleoperation and synchronized data acquisition. The three robotic actuation modules (translation, distal tip steering, and base rotation) were mapped directly to a controller, providing intuitive joystick-based control for the operator. Controller inputs were converted into motor commands via a microcontroller interface, ensuring responsive execution during catheter navigation. A graphical interface was implemented to configure system parameters such as motor speed, communication settings, and calibration routines. The interface also displayed a continuous live feed from the top-mounted camera, which served as the operator’s sole source of visual feedback. This integration of joystick control, motor interfacing, and live visualization enabled precise and consistent catheter navigation, while also allowing for the synchronized recording of motor states and video frames.

\subsection{Data Compilation}
\label{Data Compilation}
For dataset compilation, the catheter was navigated toward nine predefined target points within the vessel phantom. Each trial began from the same femoral entry site to ensure consistent initial conditions across experiments. An expert operator guided the distal tip exclusively using the live camera feed. Every navigation sequence involved advancing the catheter to a target point, and each experiment was repeated five times to ensure robustness. Only forward advancement phases were recorded, while retraction was excluded to avoid introducing inconsistent motion patterns. During all procedures, synchronized video frames and corresponding robot states were logged, resulting in a paired dataset of visual observations and control actions suitable for model training and evaluation (Please see section \ref{result} for more details).
\section{Cath-VA: A Vision-Action model for Catheter Navigation}
\label{method}
\begin{figure*}
    \centering
    \includegraphics[width=\linewidth]{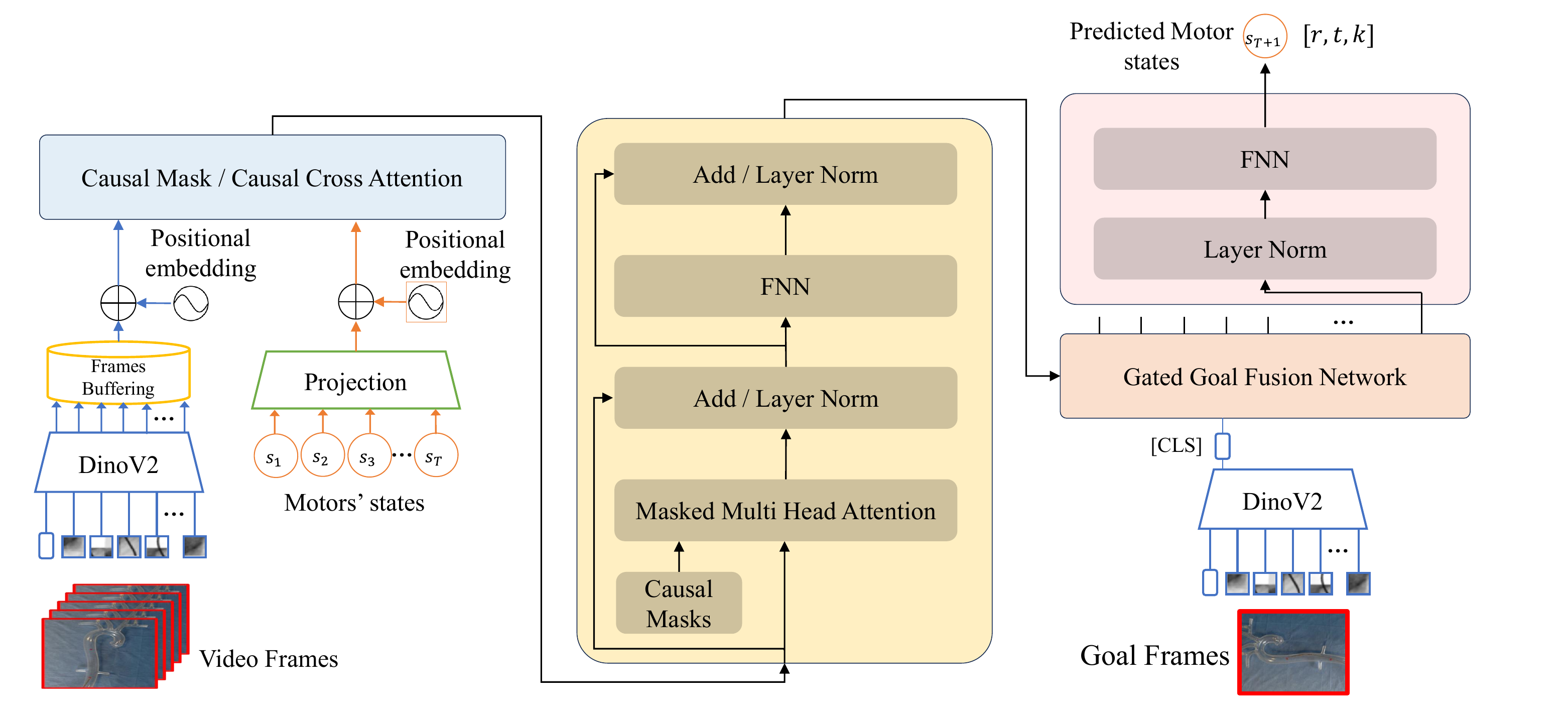}
    \caption{This figure illustrates the architecture of DINO-CVA. The model receives video frames of the catheterization task along with the motors’ states (joystick inputs). The frames are encoded using a frozen DINOv2 backbone and buffered to form a temporal sequence. The motors’ states are linearly projected to match the visual embedding space. After positional embeddings are applied, cross-attention fuses the two modalities, while spatial and temporal dependencies are modeled by a causal transformer. The fused representation is then conditioned on the target frame through a gated fusion network, and the action head predicts the next state $s_{t+1}$.}
    \label{fig:Net}
\end{figure*}
In general, autonomous catheterization systems can be understood as the integration of two tightly coupled components: a perception module and a control module. The perception module is responsible for real-time situational awareness, extracting information such as vascular geometry, catheter position, and tissue interactions from the operative scene. The control module then translates this information into precise actuation commands to achieve safe and efficient navigation. Together, these modules enable more reliable catheter steering with the potential to shorten procedure times, reduce navigation errors, and improve overall clinical outcomes. Within the control module, two core functions are traditionally emphasized: path planning and trajectory prediction. Path planning refers to the process of identifying a feasible route for the catheter to travel from its current position to the target location while respecting anatomical constraints and avoiding collisions with vessel walls. Trajectory prediction then produces the time-ordered sequence of states or control inputs required to safely execute the planned path. This work introduces a goal-conditioned behavior cloning framework that unifies perception and control within a single architecture. Instead of explicitly computing paths and generating trajectories, the model learns navigation behavior directly from expert demonstrations by mapping visual and kinematic patterns to control actions. The proposed approach is designed to be deployable on existing robotic-assisted catheter platforms, such as Sensei or Magellan, thereby extending their current follow–leader functionality toward autonomous navigation. In the following sections, the proposed methodology is described in detail.

\section{Methodology}
\label{Methodology}
The proposed model, referred to as DINO-CVA (DINO Catheter Vision-Action model), is a supervised multimodal framework designed to steer the catheter toward a given destination by predicting the controller signals of a robotic-assisted catheter platform from combined vision and action inputs. Unlike kinesthetic or purely kinematic skill modeling approaches, Dino-CVA integrates these two modalities over time, enabling the robot to be controlled by anticipating future controller states~\cite{fekri1, fekri2, farzad, farzad_2}. As a behavior cloning model trained on expert demonstrations, it further improves generalization by conditioning its output on the visual information of the final target location in the vessel. Figure~\ref{fig:Net} shows the architecture of Dino-CVA. Formally, Dino-CVA is provided with an input sequence $X={F, S}$ of length $N$, where $F$ denotes the video frames of the catheterization scene and $S$ represents the controller states of each motor in the robotic platform described above. Each video frame is represented as an image $I\in R^{h\times w \times c}$ while each controller state is denoted as $s \in R^{[r,t,k]}$ corresponding to translation, rotation, and knob movements, respectively. The images in the frame sequence are encoded by the pre-trained DinoV2, and the token embeddings are extracted~\cite{dinov2}. These tokens are buffered as follows:
\begin{align}
\label{7}
 F_{N} = \{ f_1, f_2, ..., f_N \}, f_i = \{ e_{i1}, e_{i2}, ...e_{i_P} \}
\end{align}
where $F_N \in R^{N\times P \times d}$ in which $N$ denotes the number of frames ($f_N$) in the sequence, $P$ is the number of patches per frame, and $d$ represents the embedding ($e_{ip}$) dimension of each patch, including the [CLS] token. The motors' states $S_N\in R^{N \times 3}$ are projected to match the $e_{ip}$ embedding space:

\begin{align}
\label{7}
 \hat{S}_{N} = \phi_s (S_N) = S_NW_s+b
\end{align}
Here $W_s \in R^{3 \times d}$, $b \in R^d$ and $\hat{S}_{N} \in R^{N \times d}$. Next, both modalities, the projected motor states $\hat{S}_N$ and the visual features  $F_N$, are passed through learnable positional embedding layers to preserve temporal ordering and contextual alignment \cite{vit}. 

\begin{align}
\label{7}
 \Tilde{F}_N = F_N + E_F, \Tilde{S}_N = \hat{S}_N + E_S  
\end{align}
where $E_F \in R^{N \times d}$ and $E_S \in R^{N \times d}$ are two independent set of learnable parameters. For the visual modality, the positional embedding $E_F$ is applied uniformly across all patch embeddings of a given frame. In other words, each frame in the buffered sequence receives a temporal encoding that is shared by all of its
$P$ patch embeddings, ensuring consistency in temporal alignment while preserving the spatial structure within frames. It is worth mentioning that we use learnable positional embeddings since our model operates on a fixed-length input sequence. For variable-length inputs, sinusoidal positional embeddings are typically preferred because they can handle arbitrary sequence lengths without retraining, unlike learnable embeddings. Next, the temporally augmented modalities $\Tilde{S}_N$ and $\Tilde{F}_N$ are each normalized with LayerNorm (LN) and then coupled through a masked causal cross-attention block. Let $\Tilde{S}_N \in R^{N\times d}$ denote the query sequence and let the visual tokens be $\Tilde{F}_N \in R^{N\times P \times d}$. The visual tokens are flattened along the patch axis to form the key–value sequence 
$K,V \in R^{(NP)\times d}$. A binary attention mask 
$M \in \{0,1\}^{N\times(NP)}$ enforces temporal causality by blocking access from time index
$t$ to any future frame $t'> t$, while allowing attention over all patches within frames $t'\leq t$ \cite{attentionallyouneed}. Formally,

\begin{equation}
\text{Attn}(Q,K,V; M) = \operatorname{softmax}\!\left(\frac{QK^{\top}}{\sqrt{d}} + M\right) V,
\end{equation}

\noindent with $Q = \Tilde{S}_N$. The output a Multi-Head Attention module with such a attetion mechanism is passed to a LayerNorm, yielding the fused state representation:

\begin{equation}
H = \operatorname{LN}\big(\operatorname{MHA}(Q, K, V; M)\big)
\end{equation}

\noindent where $H\in R^{N\times d}$ denotes the embedding sequence obtained from the fusion of the two modalities. This lets the controller-state queries at time $t$ attend to all visual patches from the current and past frames only, which preserves causality while leveraging high-resolution spatial context. $H$ is then processed by a causal transformer encoder to model temporal dependencies. The transformer consists of a stack of encoder layers, each comprising multi-head self-attention and feedforward submodules with \textit{GELU} activations. As mentioned before, to ensure causality, a temporal mask $M^T$ is applied as well (see Fig. \ref{fig:Net}). The output $\hat{H}\in R^{N\times d}$ is the temporally contextualized embeddings. This step allows the model to capture sequential structure in the catheter navigation process while preserving autoregressive causality. 
\par
Next, $\hat{H}$ embeddings are first normalized and then conditioned on the surgical goal. The goal representation is derived from the target image, which encodes the final destination of the catheter within the vascular system. This image is encoded using the pre-trained DINOv2 vision transformer, and only the [CLS] token is retained as the goal embedding $g \in R^d$. The [CLS] token is used because it provides a compact global summary of the image, avoiding patch-level redundancy while preserving semantic information relevant to the surgical objective. The normalized temporal embeddings $\hat{H}$ are fused with the goal embedding using a Gated Goal Fusion (GGF) module, which adaptively balances the influence of temporal dynamics and goal features. The goal embedding is broadcast across the temporal dimension and concatenated with each timestep of $\hat{H}$. A gating mechanism, parameterized by a linear projection followed by a sigmoid activation, then regulates the contribution of each modality. Formally,

\begin{equation}
G = \sigma\!\big([\hat{H} \,\|\, g] W + \mathbf{b}\big)
\end{equation}
where $[\hat{H}||g] \in R^{N \times 2d}$ denotes the concatenation of the temporal embeddings $\hat{H}$ and the goal embedding $g$, $W \in R^{2d \times d}$ and $b \in R^d$ are learnable projection parameters of the gate, and $\sigma(.)$ is the element-wise sigmoid activation that maps the output to the range (0,1). Here $G$ is the gate coefficient regulating the relative contribution of the temporal embeddings $\hat{H}$ and the goal representation $g$ and the fused representation $Z \in R^{N \times d}$:

\begin{equation}
Z = G\odot \hat{H} 
    + \Big(1 - G\Big) \odot g,
\end{equation}
where $\odot$ denotes the element-wise product. Finally, $Z$ is passed to the action head, which maps the goal-conditioned embeddings to the predicted controller states of the robotic catheter platform. The action head is implemented as a simple multilayer perceptron (MLP) composed of successive linear layers with ReLU activations. To generate the output, only the final timestep of the goal-conditioned sequence is selected $\hat{a} = MLP(Z_N)$. This design reflects the autoregressive nature of the task, as each prediction depends on the accumulated temporal context and the surgical goal. The output $\hat{a} \in R^{3}$ corresponds to the predicted controller states, representing translation, rotation, and knob movements of a sequence. The model is trained in a supervised manner by minimizing the mean squared error (MSE) loss between the predicted states $\hat{a}$ and the ground truth states $s_{t+1}$.

\section{Results and Discussions}
\label{result}
\begin{table}[t]
\centering
\caption{Data split statistics for scenario-based and episode-based datasets.}
\label{tab:split_stats_onecol}
\scriptsize
\setlength{\tabcolsep}{3pt}
\begin{tabular}{l|r|r|l|l|r|r|r|r}
\toprule
\textbf{Type} & \textbf{Frames} & \textbf{Eps} & \textbf{Points} & \textbf{Var} & \textbf{Mean} & \textbf{Std} & \textbf{Min} & \textbf{Max} \\
\midrule
\multicolumn{9}{c}{\textbf{Train}} \\
\midrule
\multirow{3}{*}{Scenario} & \multirow{3}{*}{16{,}425} & \multirow{3}{*}{25} & \multirow{3}{*}{P1-P4, P7}
& translation & 0.63 & 0.46 & -1.0 & 1.0 \\
& & & & rotation    & -0.05 & 0.26 & -1.0 & 1.0 \\
& & & & knob        & -0.08 & 0.50 & -1.0 & 1.0 \\
\cmidrule{1-9}
\multirow{3}{*}{Episode} & \multirow{3}{*}{15{,}705} & \multirow{3}{*}{27} & \multirow{3}{*}{P1--P9}
& translation & 0.68 & 0.45 & -0.18 & 1.0 \\
& & & & rotation    & -0.05 & 0.24 & -1.0 & 1.0 \\
& & & & knob        & -0.09 & 0.47 & -1.0 & 1.0 \\
\midrule
\multicolumn{9}{c}{\textbf{Val}} \\
\midrule
\multirow{3}{*}{Scenario} & \multirow{3}{*}{2{,}680} & \multirow{3}{*}{5} & \multirow{3}{*}{P5}
& translation & 0.78 & 0.40 & -0.06 & 1.0 \\
& & & & rotation    & -0.01 & 0.17 & -1.0 & 0.85 \\
& & & & knob        & -0.06 & 0.34 & -1.0 & 1.0 \\
\cmidrule{1-9}
\multirow{3}{*}{Episode} & \multirow{3}{*}{5{,}276} & \multirow{3}{*}{9} & \multirow{3}{*}{P1--P9}
& translation & 0.67 & 0.45 & -1.0 & 1.0 \\
& & & & rotation    & -0.06 & 0.22 & -1.0 & 1.0 \\
& & & & knob        & -0.09 & 0.47 & -1.0 & 1.0 \\
\midrule
\multicolumn{9}{c}{\textbf{Test}} \\
\midrule
\multirow{3}{*}{Scenario} & \multirow{3}{*}{7{,}139} & \multirow{3}{*}{15} & \multirow{3}{*}{P6, P8, P9}
& translation & 0.75 & 0.42 & -0.18 & 1.0 \\
& & & & rotation    & -0.06 & 0.22 & -1.0 & 0.77 \\
& & & & knob        & -0.13 & 0.40 & -1.0 & 1.0 \\
\cmidrule{1-9}
\multirow{3}{*}{Episode} & \multirow{3}{*}{5{,}263} & \multirow{3}{*}{9} & \multirow{3}{*}{P1--P9}
& translation & 0.67 & 0.46 & -1.0 & 1.0 \\
& & & & rotation    & -0.05 & 0.26 & -1.0 & 1.0 \\
& & & & knob        & -0.10 & 0.45 & -1.0 & 1.0 \\
\bottomrule
\end{tabular}
\end{table}
This section presents the procedures undertaken for data preparation, including the construction of training, validation, and test sets. The model configuration and training setup are then described in detail. Following this, the evaluation results are reported and critically analyzed, with particular emphasis on the ablation studies conducted to assess the contribution of different components of the proposed approach.

\begin{figure*}[t]
\centering
\subfloat[Point\_1]{\includegraphics[width=0.29\textwidth]{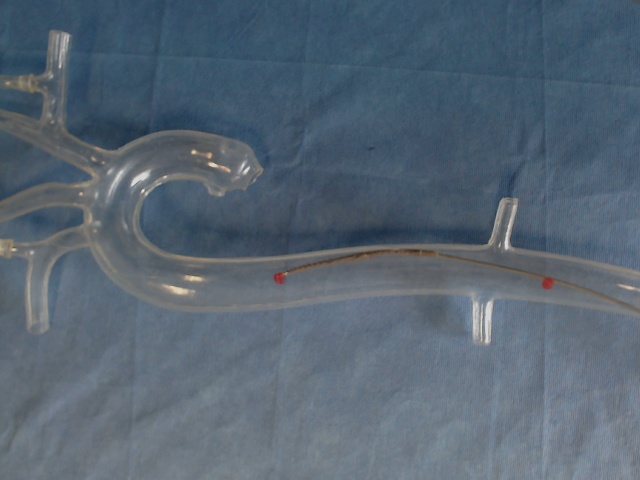}}\hspace{0.5em}
\subfloat[Point\_2]{\includegraphics[width=0.29\textwidth]{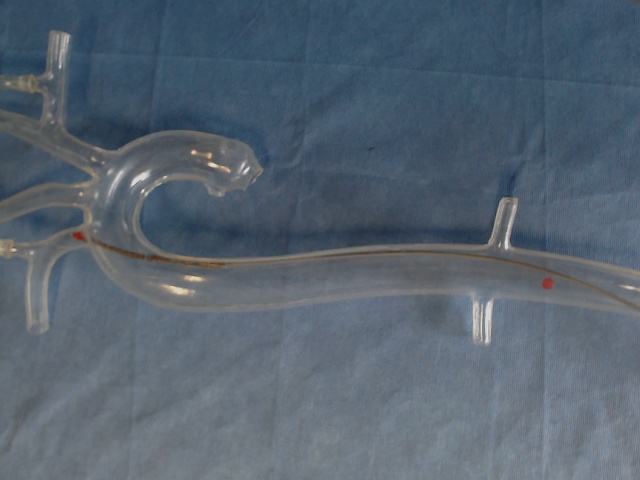}}\hspace{0.5em}
\subfloat[Point\_3]{\includegraphics[width=0.29\textwidth]{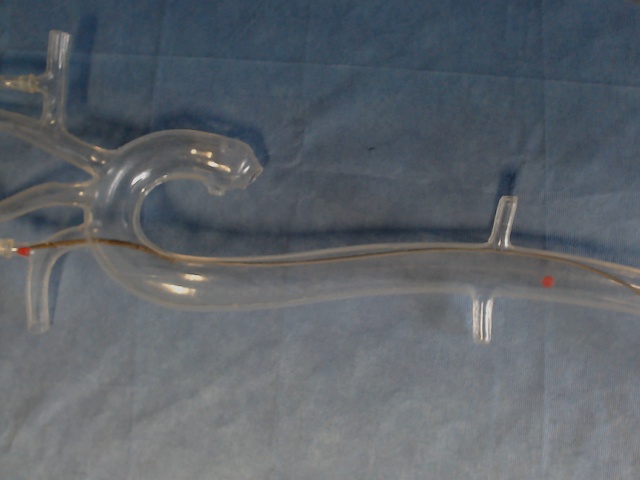}}\\[1ex]

\subfloat[Point\_4]{\includegraphics[width=0.29\textwidth]{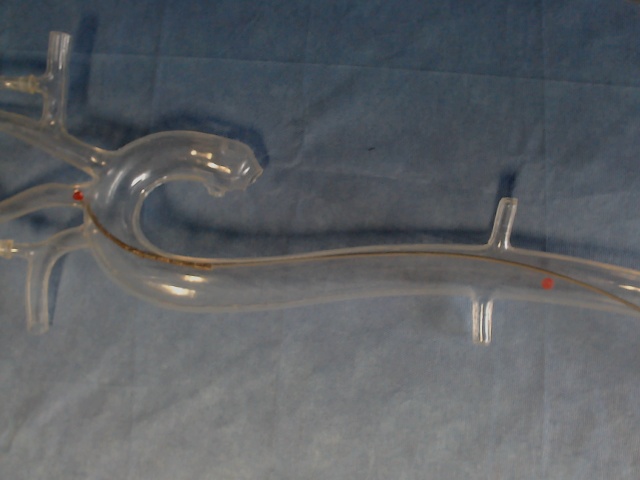}}\hspace{0.5em}
\subfloat[Point\_5]{\includegraphics[width=0.29\textwidth]{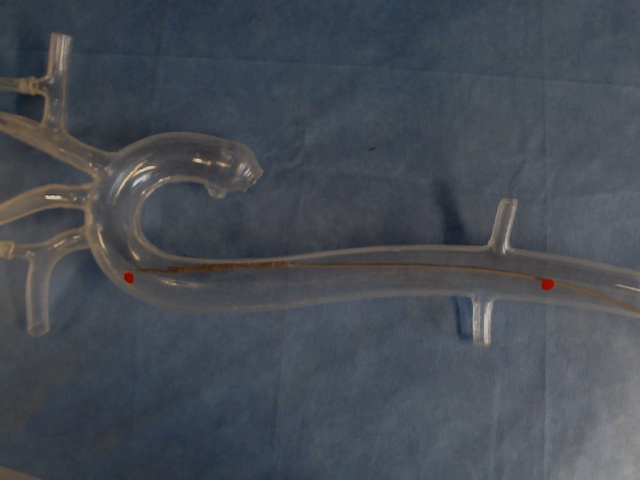}}\hspace{0.5em}
\subfloat[Point\_6]{\includegraphics[width=0.29\textwidth]{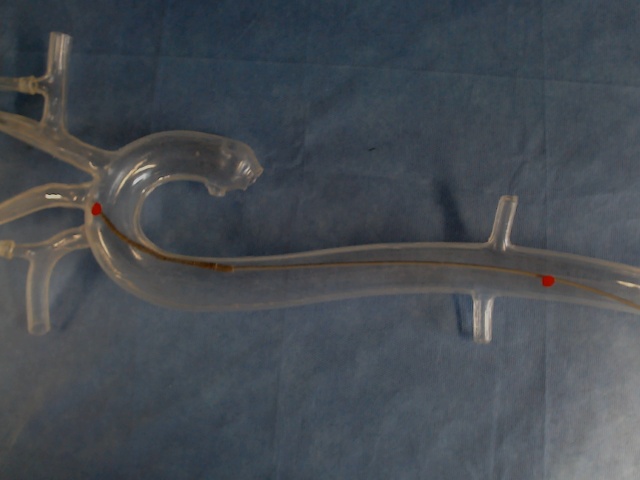}}\\[1ex]

\subfloat[Point\_7]{\includegraphics[width=0.29\textwidth]{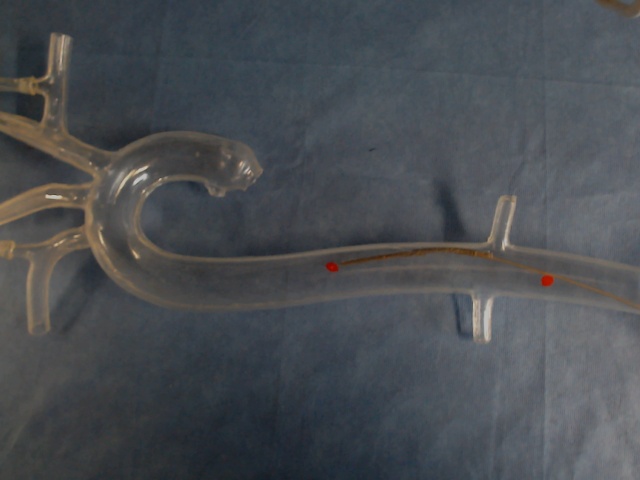}}\hspace{0.5em}
\subfloat[Point\_8]{\includegraphics[width=0.29\textwidth]{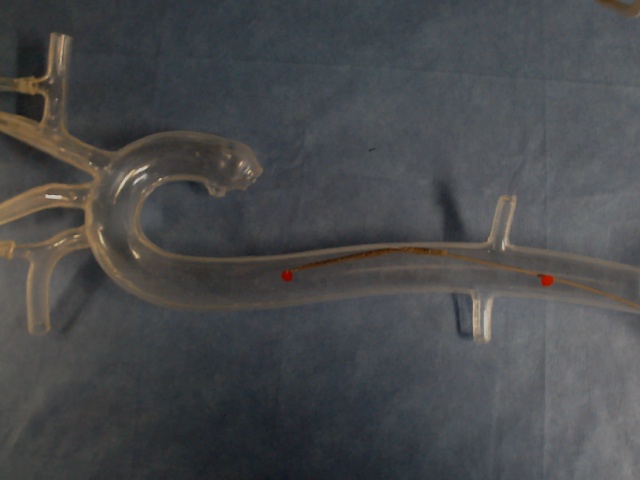}}\hspace{0.5em}
\subfloat[Point\_9]{\includegraphics[width=0.29\textwidth]{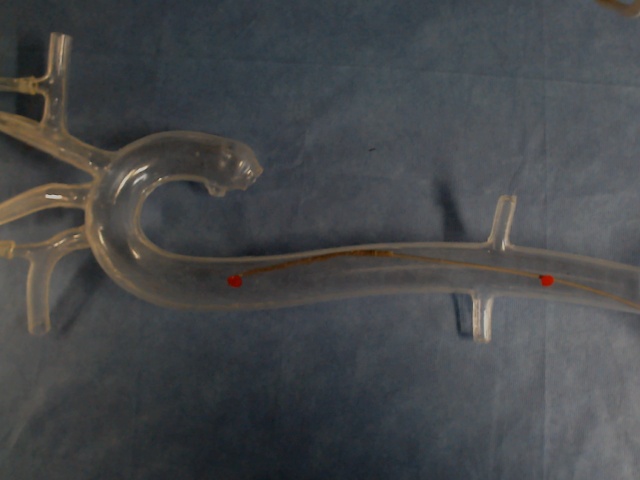}}
\caption{The figure depicts nine predefined starting and ending points, indicated by red dots on the artificial vasculature. These images also represent sample visual data used to train and evaluate DINO-CVA. The starting point was kept identical across all experiments, while the target endpoints varied among the nine locations.}
\label{fig:nine-row}
\vspace{-3 mm}
\end{figure*}


\subsection{Data Preparation and Model Configuration}
\label{Data Preparation and Model Configuration}
\noindent \textbf{Dataset Preparation:} As outlined in Section III, the datasets for this study were compiled using the experimental setup in which invited participants performed catheter steering tasks. Specifically, nine distinct tasks were defined, each characterized by a designated starting and ending point marked with two red dots on the phantom artificial veins (see Fig. \ref{fig:nine-row}). During each trial, the operator navigated the catheter while aiming to maintain its trajectory near the vein centerline and minimize collisions with the vessel walls as much as possible. The custom-developed software (\ref{Interface Software}) simultaneously captured top-view video recordings and logged the joystick states. Each task was repeated five times, resulting in a dataset where each data point consists of an RGB frame $I \in \mathbb{R}^{224 \times 224 \times 3}$ and a state vector $s \in \mathbb{R}^{3}$ representing translation, rotation, and knob movement. For clarity, each defined task is referred to as a scenario, and each repetition of a scenario is referred to as an episode. Based on the compiled dataset, two distinct training and validation protocols were established:

\begin{itemize}
    \item \textbf{Episode-based:} In this setting, the training, validation, and test sets include episodes drawn from all scenarios. Consequently, while the model is exposed to every scenario during training, the individual episodes may differ across runs, providing variability in execution.  

    \item \textbf{Scenario-based:} In this setting, the training, validation, and test sets do not share scenarios. Accordingly, the model is evaluated on scenarios in the test and validation sets that were entirely unseen during training, allowing assessment of its generalization capability across tasks.  
\end{itemize}
\begin{figure*}[t]
    \centering
    \subfloat[Scenario-Train]{\includegraphics[width=.32\textwidth]{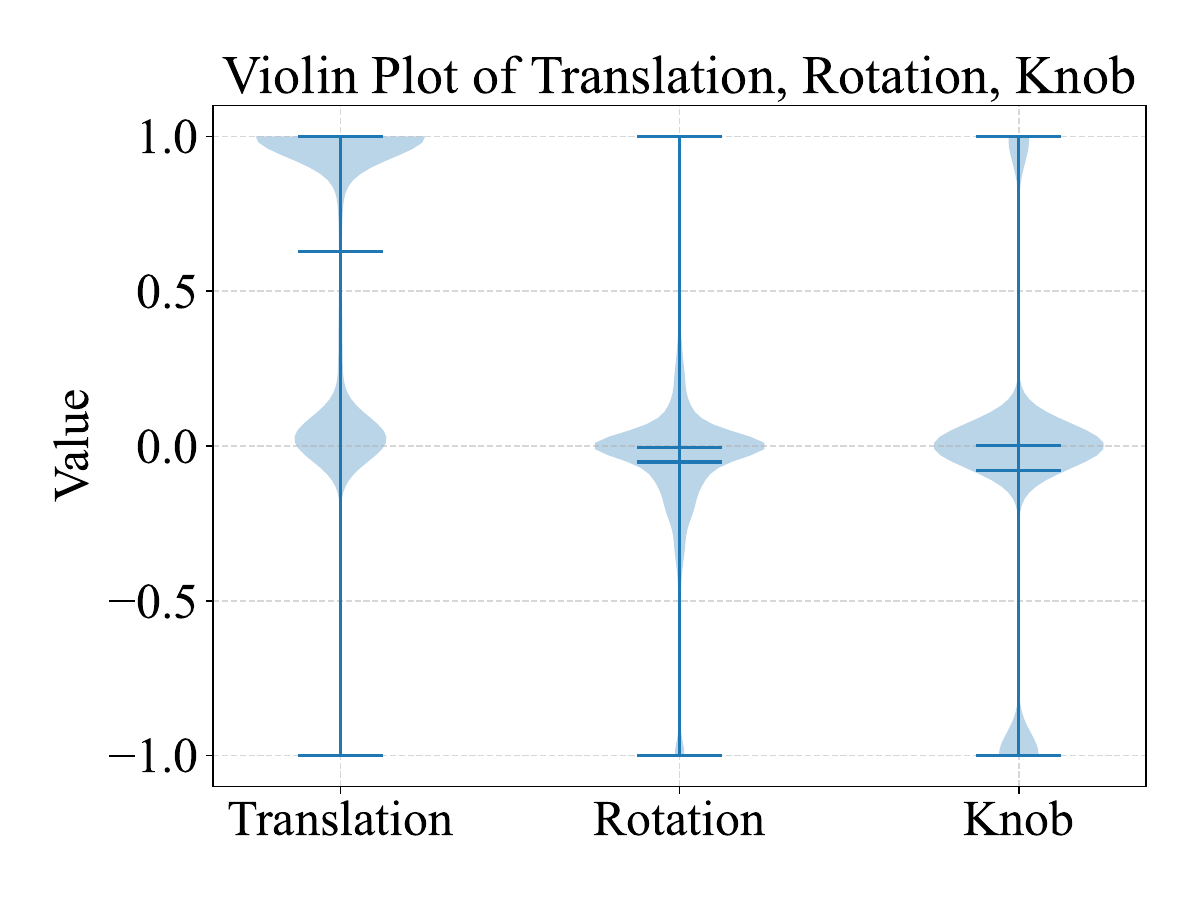}
        \label{fig:sub1}}
    \hspace{-2mm}
    \subfloat[Scenario-Test]{\includegraphics[width=.32\textwidth]{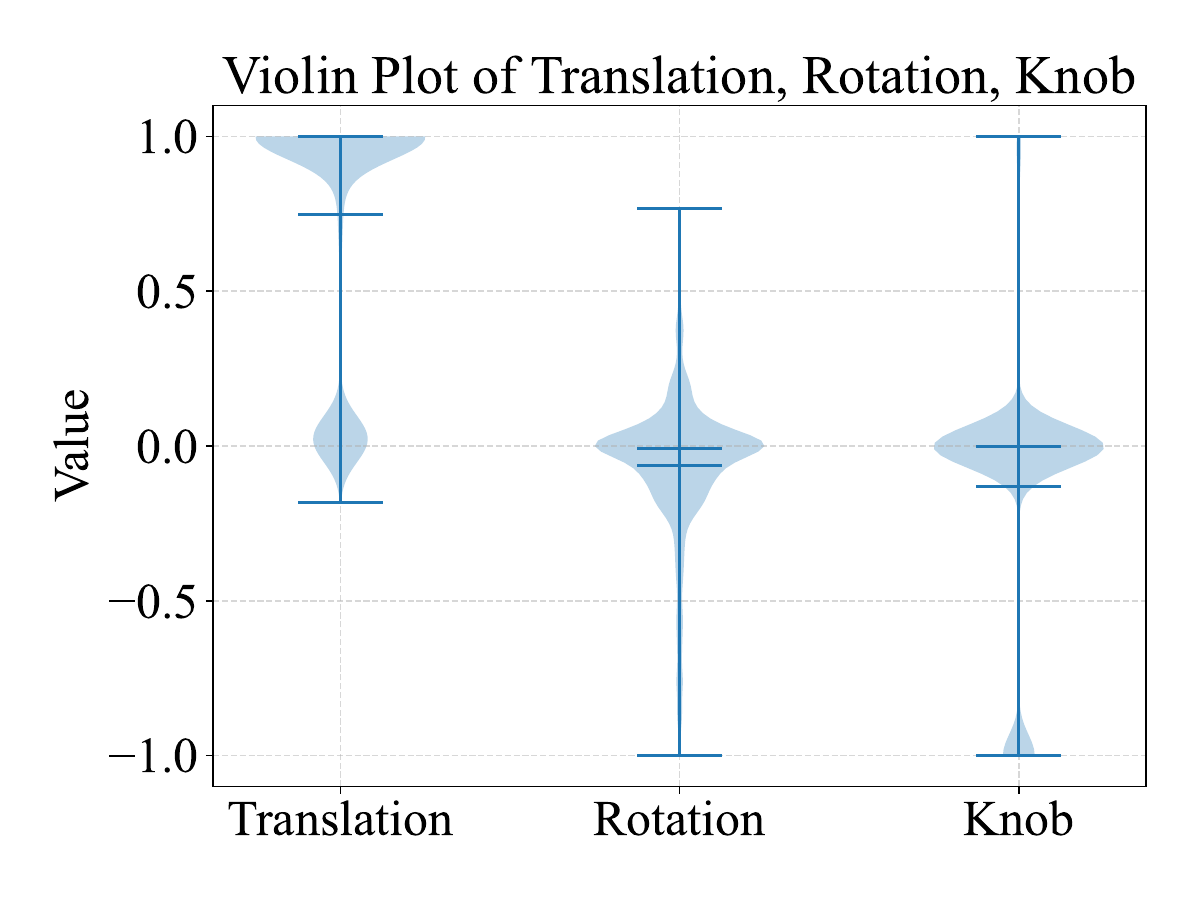}
        \label{fig:sub2}}
    \hspace{-2mm}
    \subfloat[Scenario-Val]{\includegraphics[width=.32\textwidth]{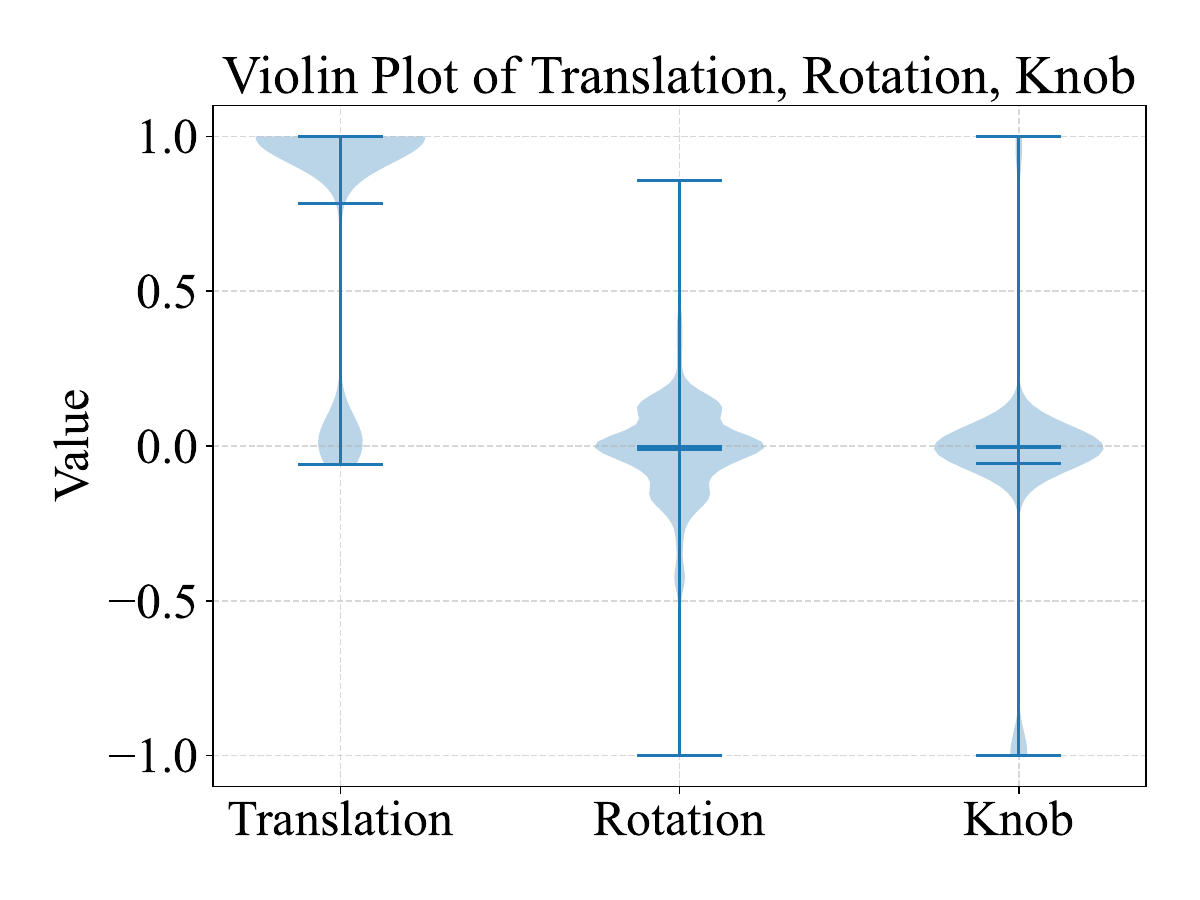}
        \label{fig:sub3}}
        
    \subfloat[Episode-Train]{\includegraphics[width=.32\textwidth]{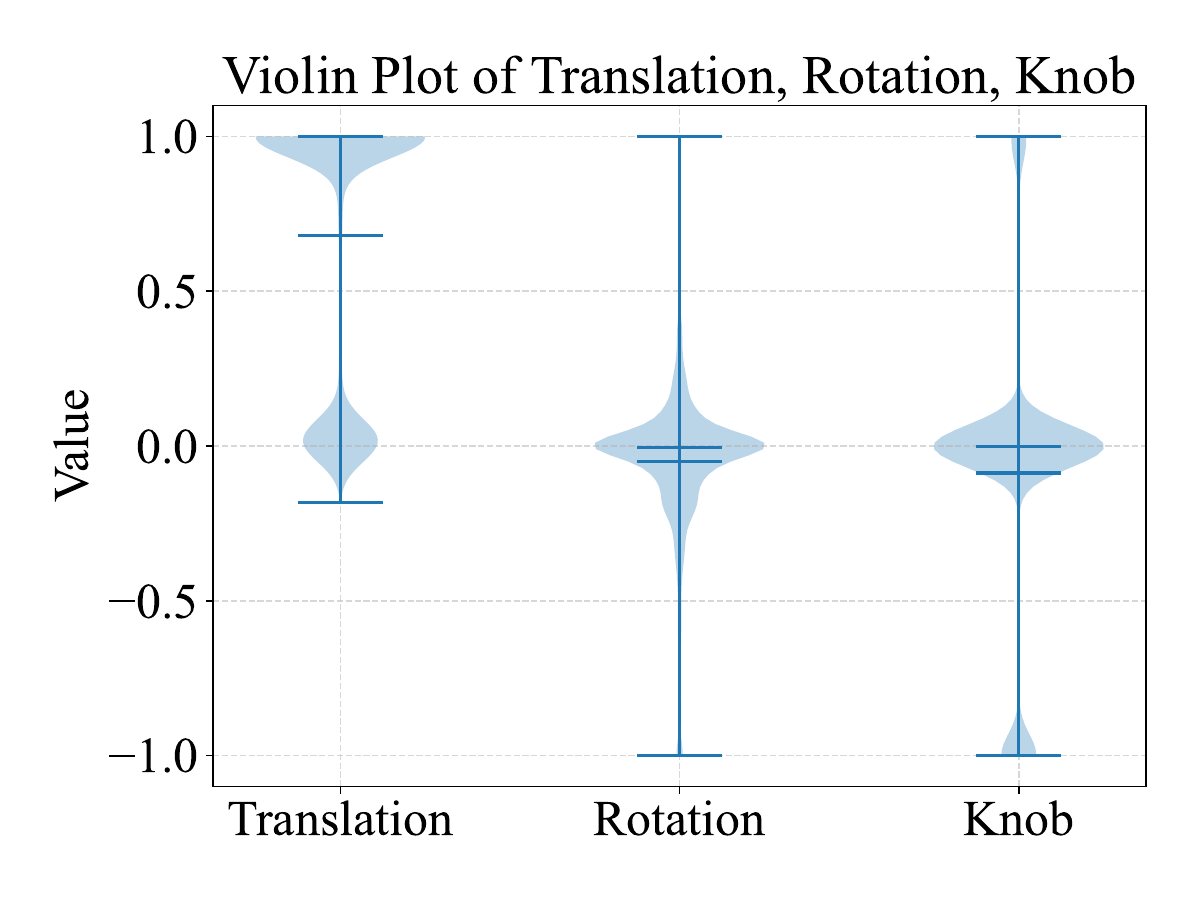}
        \label{fig:sub5}}
    \hspace{-2mm}
    \subfloat[Episode-Test]{\includegraphics[width=.32\textwidth]{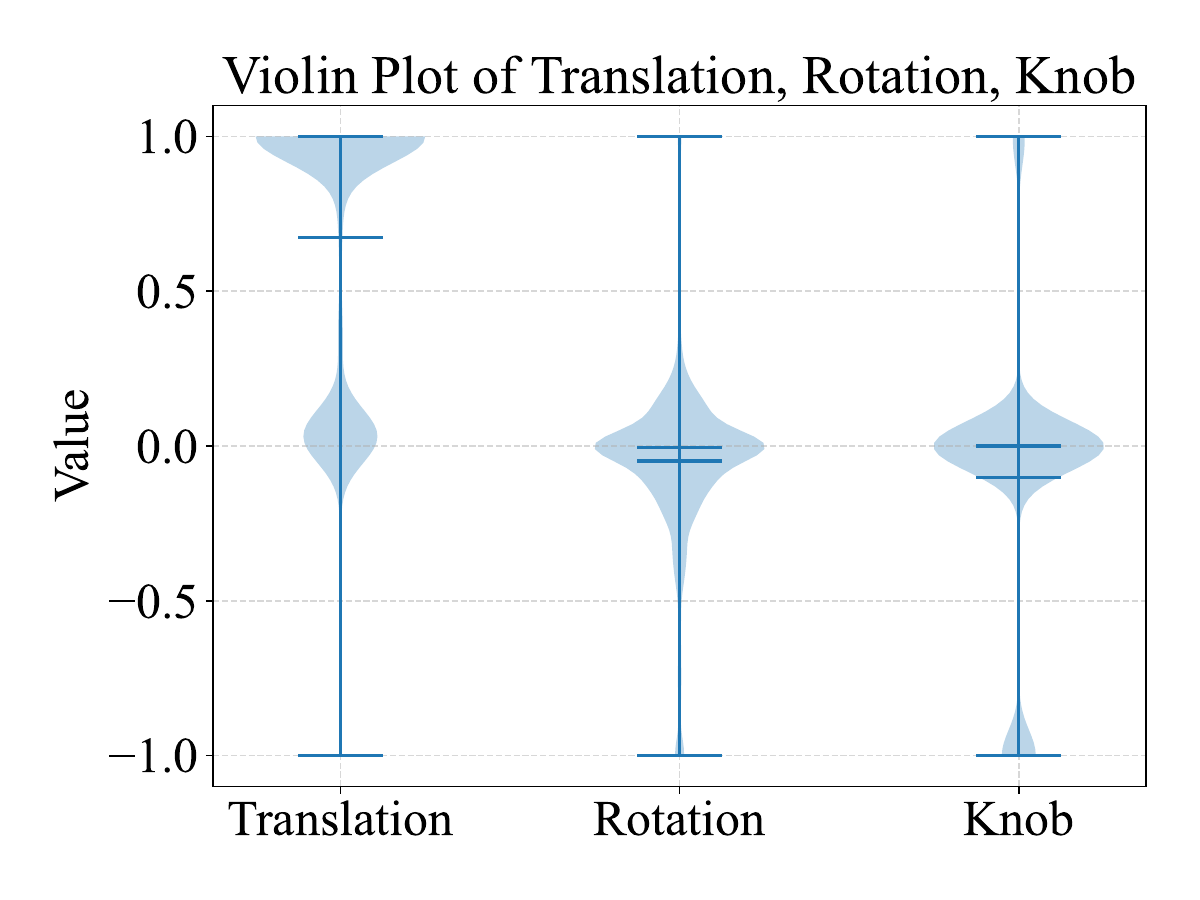}
        \label{fig:sub6}}
    \hspace{-2mm}
    \subfloat[Episode-Val]{\includegraphics[width=.32\textwidth]{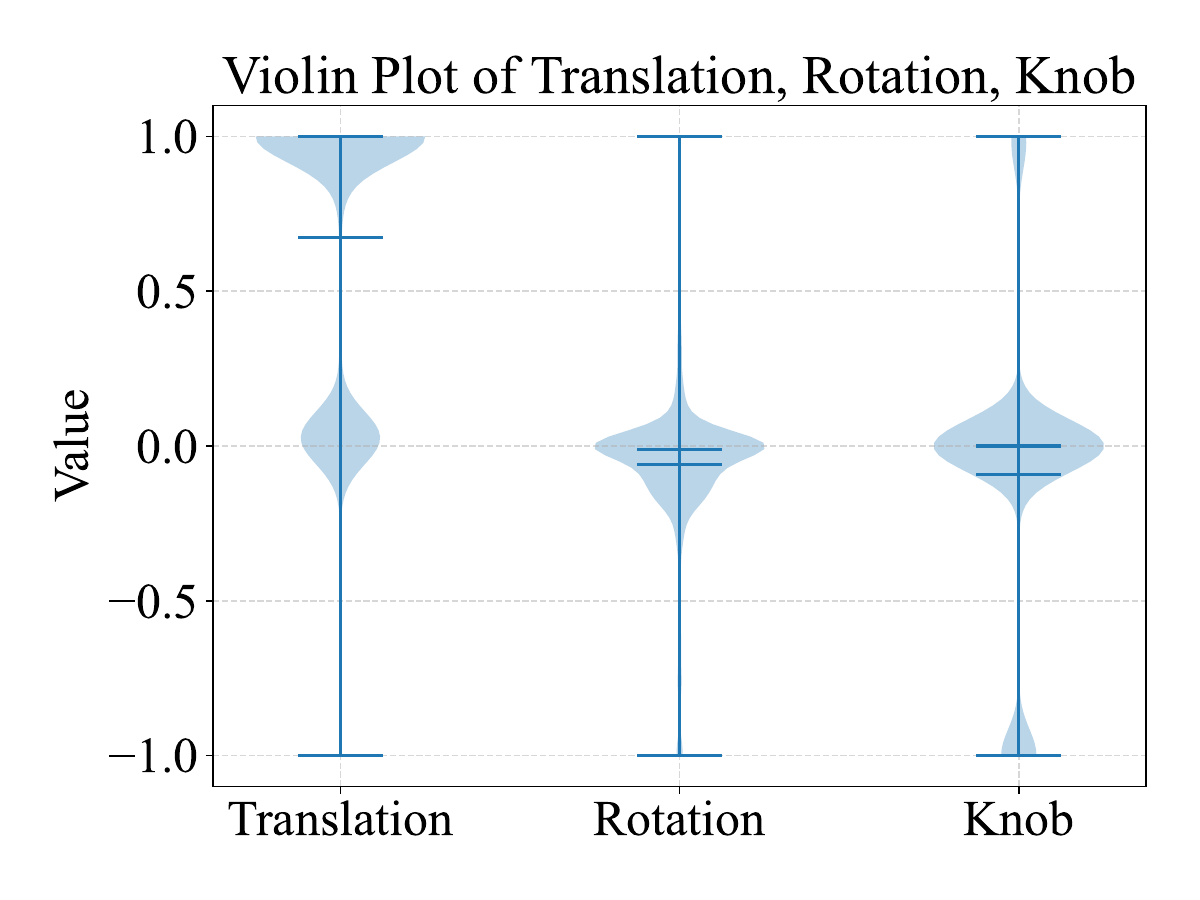}
        \label{fig:sub7}}
    \caption{Violin plots illustrating the empirical distribution of motor states across both dataset versions (episode-based and scenario-based) for training, validation, and test splits. The plots highlight the variability in catheter control inputs and their corresponding motion trajectories.}
    \label{fig:data}
\end{figure*}
Table I summarizes the dataset statistics across states, scenarios, and episodes, reflecting the two partitioning strategies. For instance, under the scenario-based setting, the training set comprises 16,425 data points from scenarios P1, P4, and P7, whereas in the episode-based setting, data points are drawn from all nine scenarios (P1–P9). Fig.~\ref{fig:data} illustrates the empirical distribution of motor states across both dataset versions, highlighting the variability in catheter control inputs and corresponding motion trajectories. To incorporate the temporal structure of the data, a sliding-window approach was employed. Each input sequence consisted of 50 consecutive time steps, which served as input to predict the subsequent step. The sequence length of 50 was selected as a practical trade-off, constrained primarily by the available hardware resources. At each timestep $t$, the input RGB frame $I_t$ is normalized using the mean and standard deviation statistics from the dataset on which DINOv2 was trained, ensuring compatibility with its pretrained feature representations~\cite{dinov2}. The motor state vector $s_t$ is standardized via z-scoring based on the training-set statistics of our dataset. The prediction target $s_{t+1}$ remains in its original physical units to preserve interpretability and avoid post-hoc rescaling at inference.
\par
\noindent \textbf{Model and Training Configuration:} DINO-CVA processes two input modalities: image sequences and joystick states. The image sequence $F_N$ is encoded into 257 patch tokens ($P=257$), each mapped to a 384-dimensional embedding space ($d=384$) using a frozen small-DINOv2 encoder. In parallel, the joystick state vectors $S_N$ are linearly projected into the same 384-dimensional feature space to ensure modality alignment $\hat{S}_N$. The input sequence $N=50$, resulting in a total of $12850$ patch tokens (i.e., $257 \times 50$). To incorporate temporal dependencies, two distinct sets of learnable positional embeddings are defined: one for the video frames and one for the joystick states. For each frame $f_i$, the positional embedding corresponding to its temporal index is applied uniformly across all 257 patch tokens, thereby encoding both spatial and temporal order within the sequence. Next, a single-layer masked multi-head cross-attention module with 8 heads is employed, where the encoded image features $\hat{F}$ serve as keys and values, and the projected joystick states $\hat{S}$ act as queries to achieve modality fusion (see Equation 4). The resulting fused embeddings are subsequently passed to a causal transformer composed of 4 layers, each equipped with 8 attention heads and a feed-forward network of dimension 1024. This configuration enables the model to capture long-range temporal dependencies while preserving the autoregressive structure required for sequence prediction. Subsequently, the GGF module incorporates the goal information by conditioning the sequence representation $\hat{H}$ with the [CLS] token of the goal image, denoted as $g \in \mathbb{R}^{384 \times 1}$, extracted from the frozen DINOv2 encoder (see Equations (6) and (7)). The final representation $Z_N$ is then forwarded to a multilayer perceptron (MLP) consisting of four fully connected layers with dimensions 512, 256, 128, and 3, respectively, to regress the next-step joystick state $s_{t+1}$.
\par The network with the aforementioned configuration was trained separately under the two dataset settings: episode-based and scenario-based. Training was performed with a batch size of 16 for up to 50 epochs, with validation conducted after each epoch. The model was optimized using the Adam optimizer with an initial learning rate of $9.0 \times 10^{-5}$, regulated by a cosine annealing scheduler. The MSE loss function was employed, and early stopping was applied based on validation performance to prevent overfitting. The implementation was carried out in PyTorch on an NVIDIA RTX 2080 GPU with 8 GB of memory. The model contained a total of 22.06M parameters, of which 6.82M were trainable.


\begin{table}[t]
\centering
\caption{Performance comparison of the proposed model and LSTM baseline on episode-based and scenario-based datasets.}
\label{tab:results}
\begin{tabular}{llcccc}
\toprule
\multicolumn{6}{c}{\textbf{Episode-based dataset}} \\
\midrule
Model & Task & MSE & RMSE & MAE & $R^2$ \\
\midrule
\multirow{3}{*}{Ours} 
  & Translation & 0.0179  & 0.1338  & 0.0472  & 0.9145 \\
  & Rotation    & 0.0091  & 0.0955  & 0.0356  & 0.8676 \\
  & Knob        & 0.0094  & 0.0970  & 0.0279  & 0.9556 \\
\midrule
\multirow{3}{*}{LSTM-based} 
  & Translation & 0.0160  & 0.1264  & 0.0448  & 0.9237 \\
  & Rotation    & 0.0085  & 0.0922  & 0.0335  & 0.8765 \\
  & Knob        & 0.0115  & 0.1075  & 0.0311  & 0.9455 \\
\midrule
\multicolumn{6}{c}{\textbf{Scenario-based dataset}} \\
\midrule
Model & Task & MSE & RMSE & MAE & $R^2$ \\
\midrule
\multirow{3}{*}{Ours} 
  & Translation & 0.0116  & 0.1078  & 0.0369  & 0.9320 \\
  & Rotation    & 0.0050  & 0.0708  & 0.0354  & 0.8858 \\
  & Knob        & 0.0187  & 0.1367  & 0.0490  & 0.9116 \\
\midrule
\multirow{3}{*}{LSTM-based} 
  & Translation & 0.0125  & 0.1120  & 0.0416  & 0.9266 \\
  & Rotation    & 0.0050  & 0.0708  & 0.0311  & 0.8859 \\
  & Knob        & 0.0133  & 0.1154  & 0.0289  & 0.9158 \\
\bottomrule
\end{tabular}
\end{table}

\begin{figure*}
\centering
\subfloat[Knob (True vs Predicted)]{\includegraphics[width=0.30\textwidth]{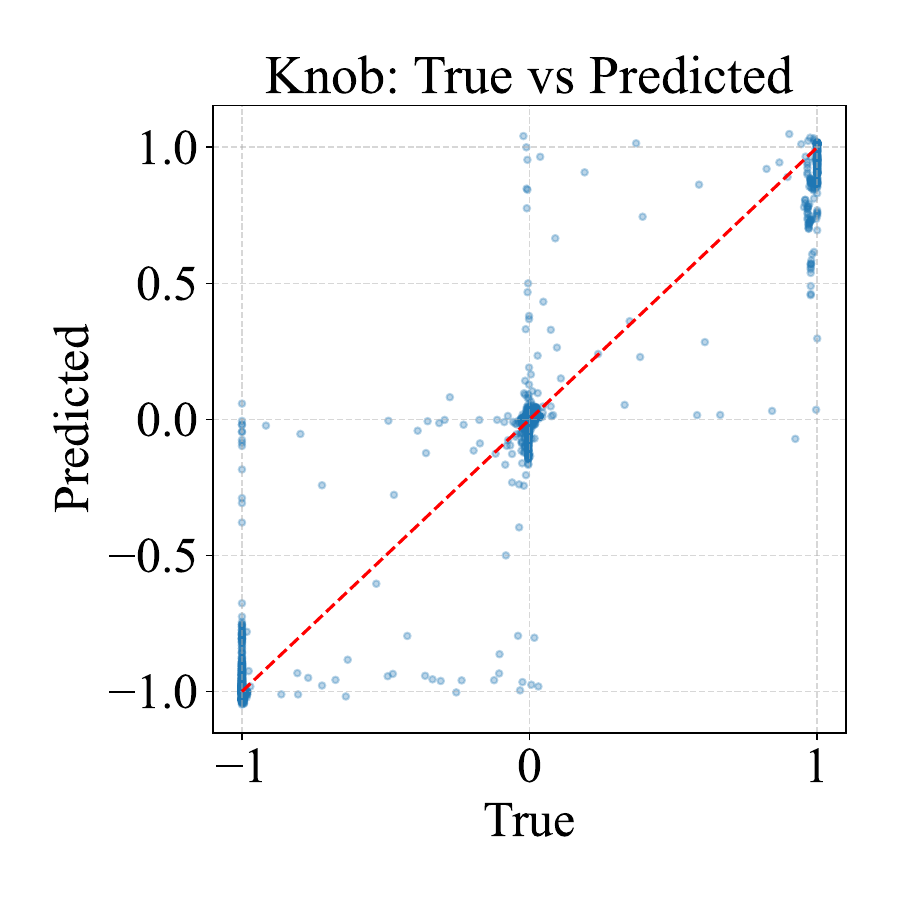}\label{fig:ep_tvsp_knob}}
\hspace{0.01\textwidth}
\subfloat[Rotation (True vs Predicted)]{\includegraphics[width=0.30\textwidth]{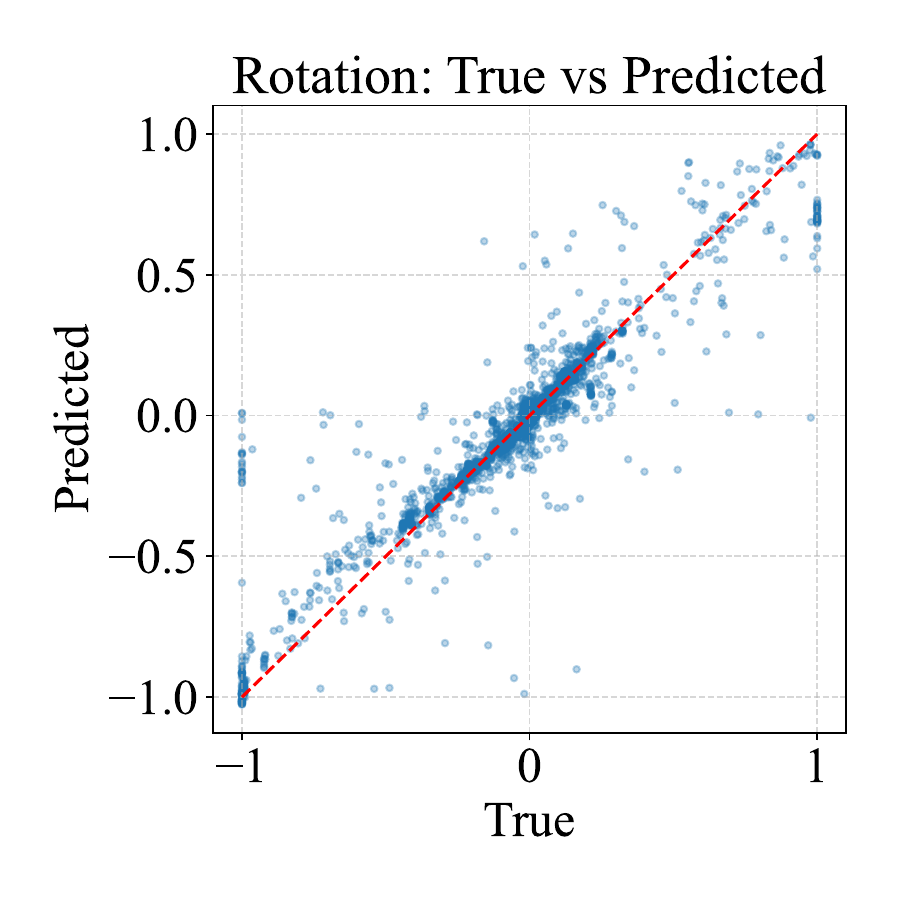}\label{fig:ep_tvsp_rot}}
\hspace{0.01\textwidth}
\subfloat[Translation (True vs Predicted)]{\includegraphics[width=0.30\textwidth]{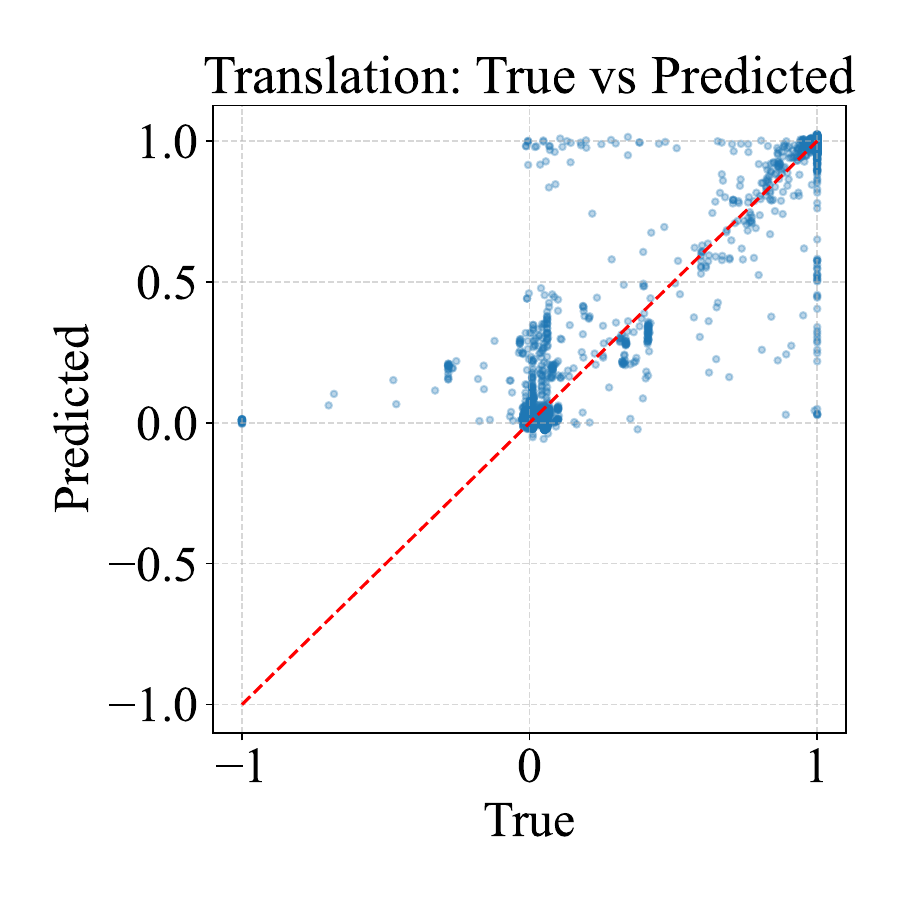}\label{fig:ep_tvsp_trans}}\\[0.8ex]

\subfloat[Knob Error Distribution]{\includegraphics[width=0.30\textwidth]{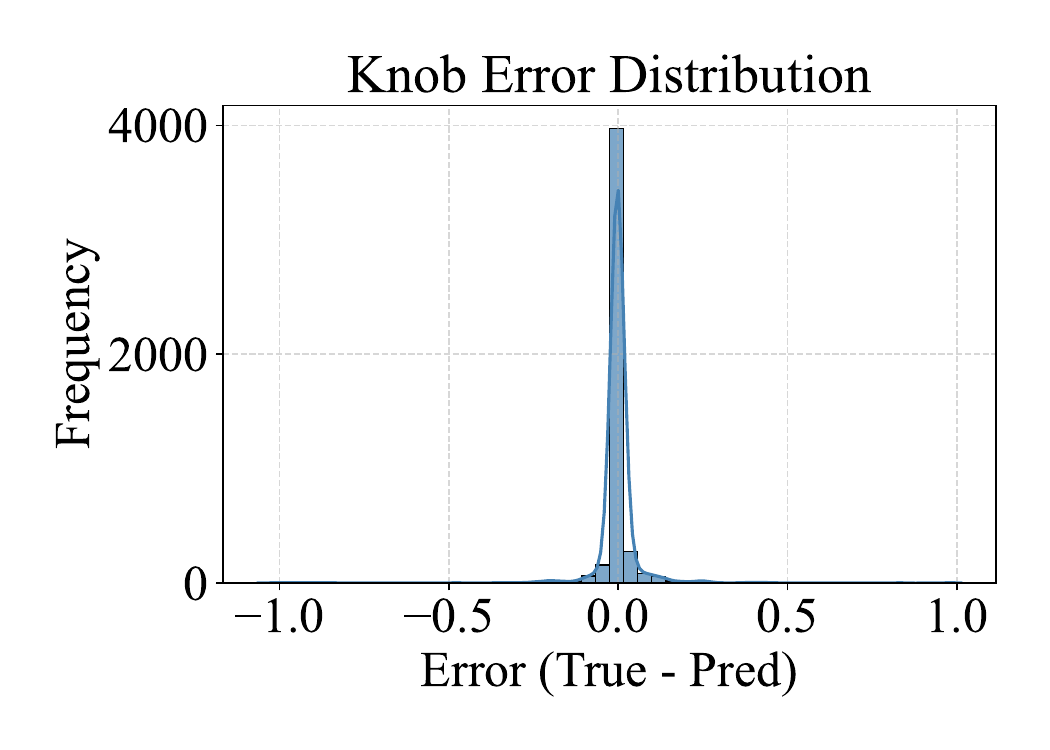}\label{fig:ep_hist_knob}}
\hspace{0.01\textwidth}
\subfloat[Rotation Error Distribution]{\includegraphics[width=0.30\textwidth]{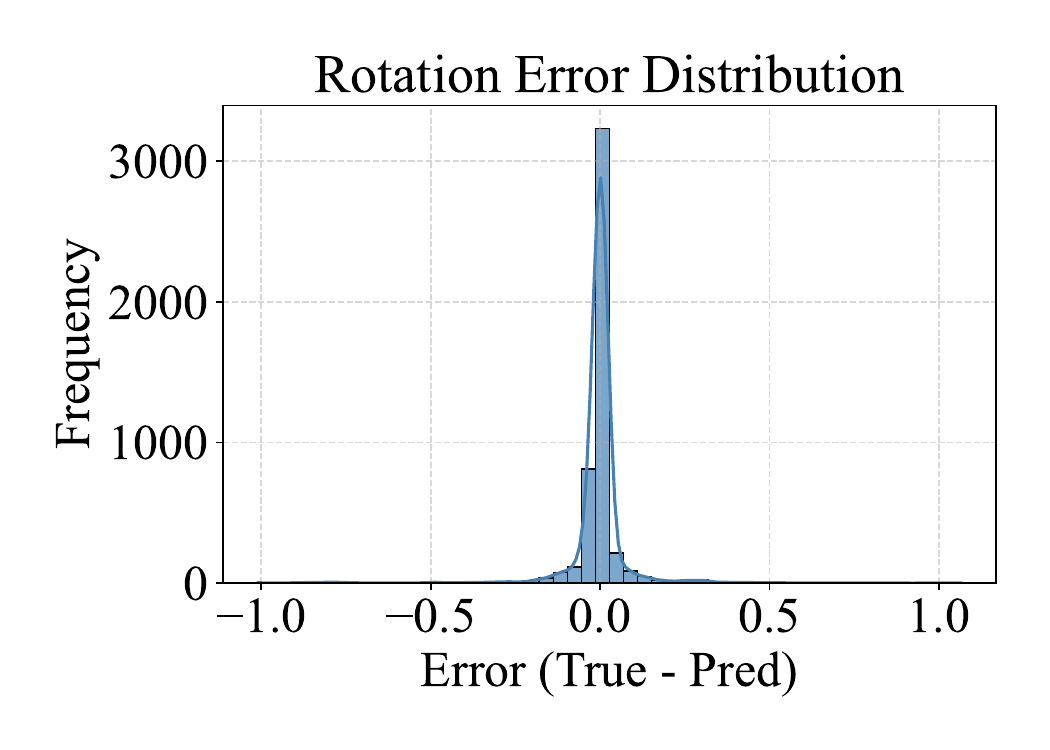}\label{fig:ep_hist_rot}}
\hspace{0.01\textwidth}
\subfloat[Translation Error Distribution]{\includegraphics[width=0.30\textwidth]{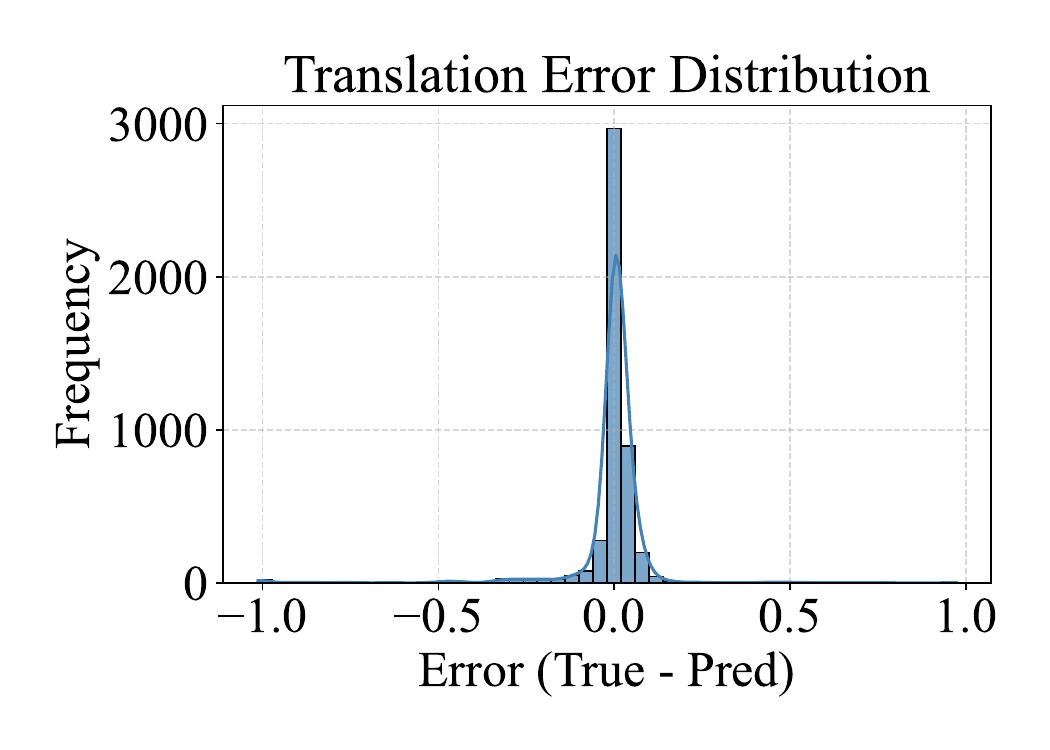}\label{fig:ep_hist_trans}}\\[1.2ex]

\subfloat[Knob (True vs Predicted)]{\includegraphics[width=0.30\textwidth]{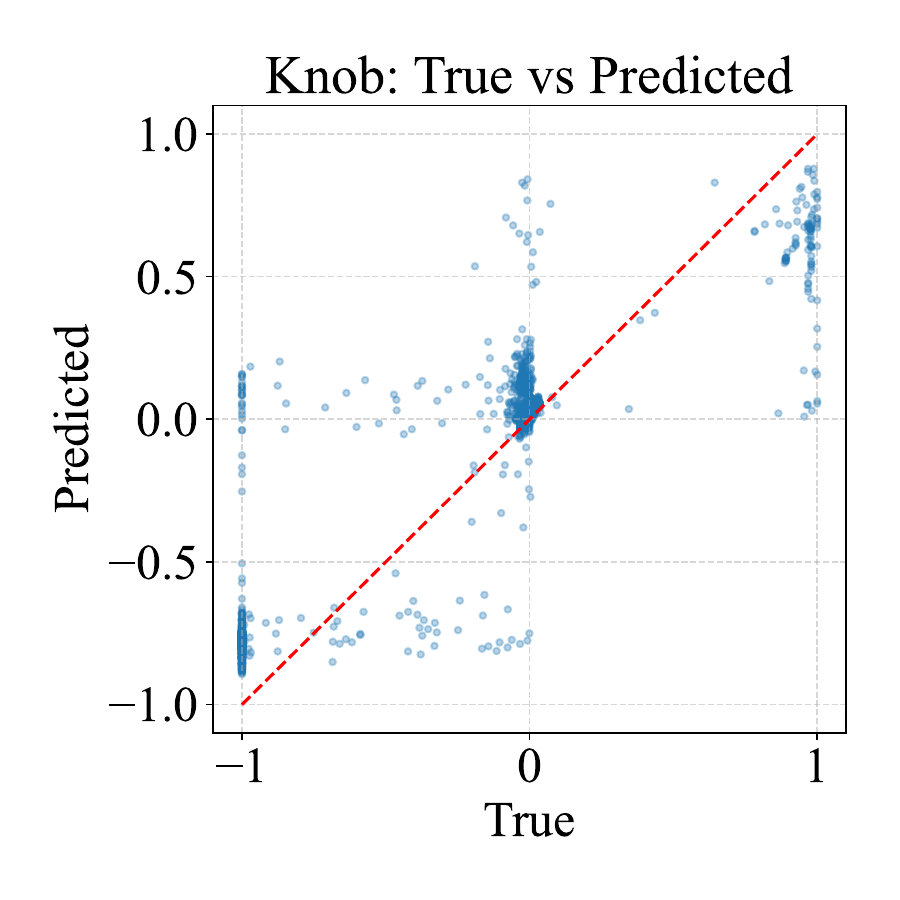}\label{fig:sc_tvsp_knob}}
\hspace{0.01\textwidth}
\subfloat[Rotation (True vs Predicted)]{\includegraphics[width=0.30\textwidth]{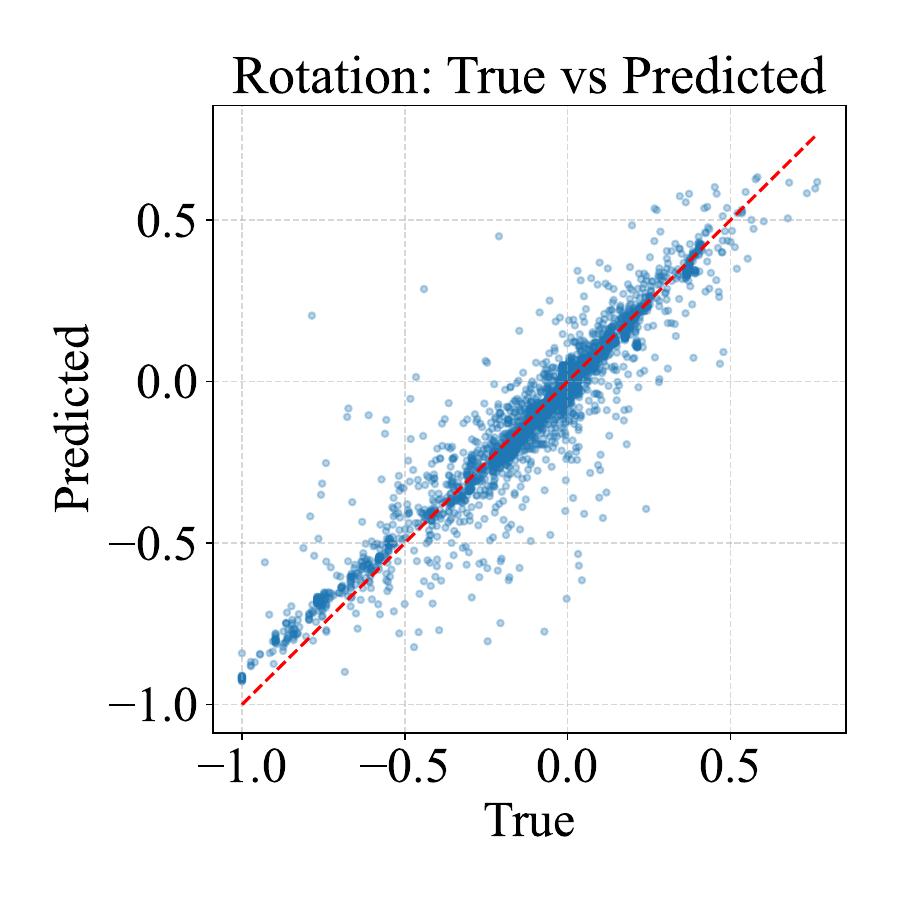}\label{fig:sc_tvsp_rot}}
\hspace{0.01\textwidth}
\subfloat[Translation (True vs Predicted)]{\includegraphics[width=0.30\textwidth]{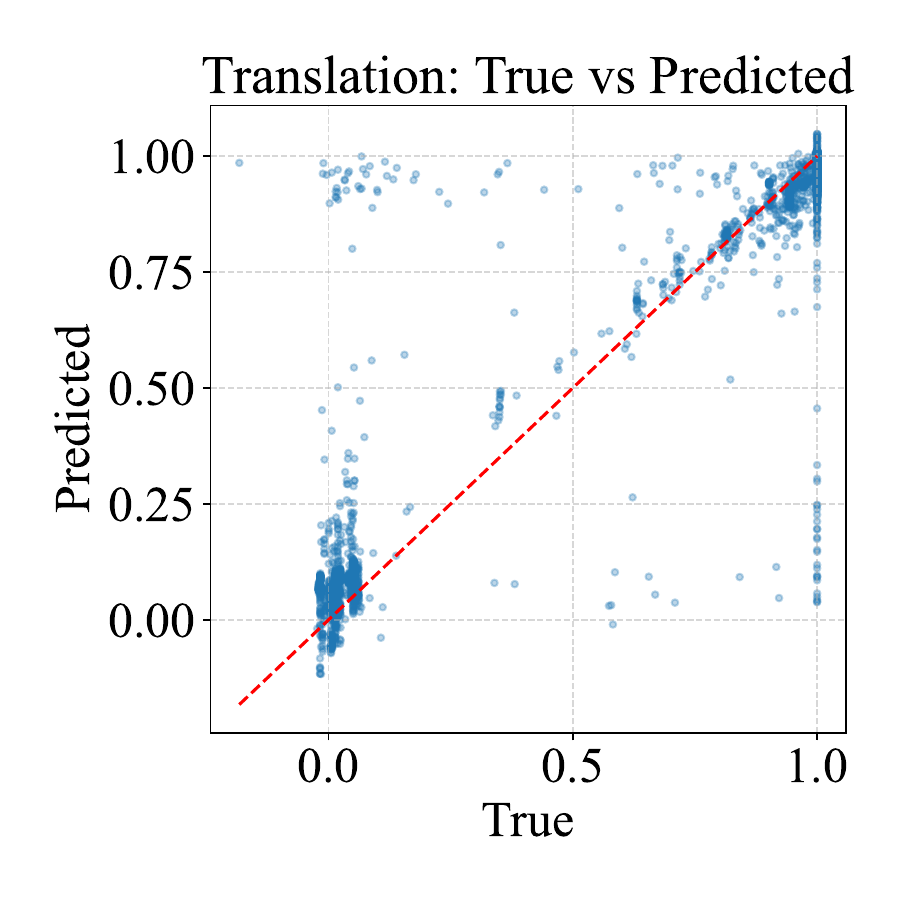}\label{fig:sc_tvsp_trans}}\\[0.8ex]

\subfloat[Knob Error Distribution]{\includegraphics[width=0.30\textwidth]{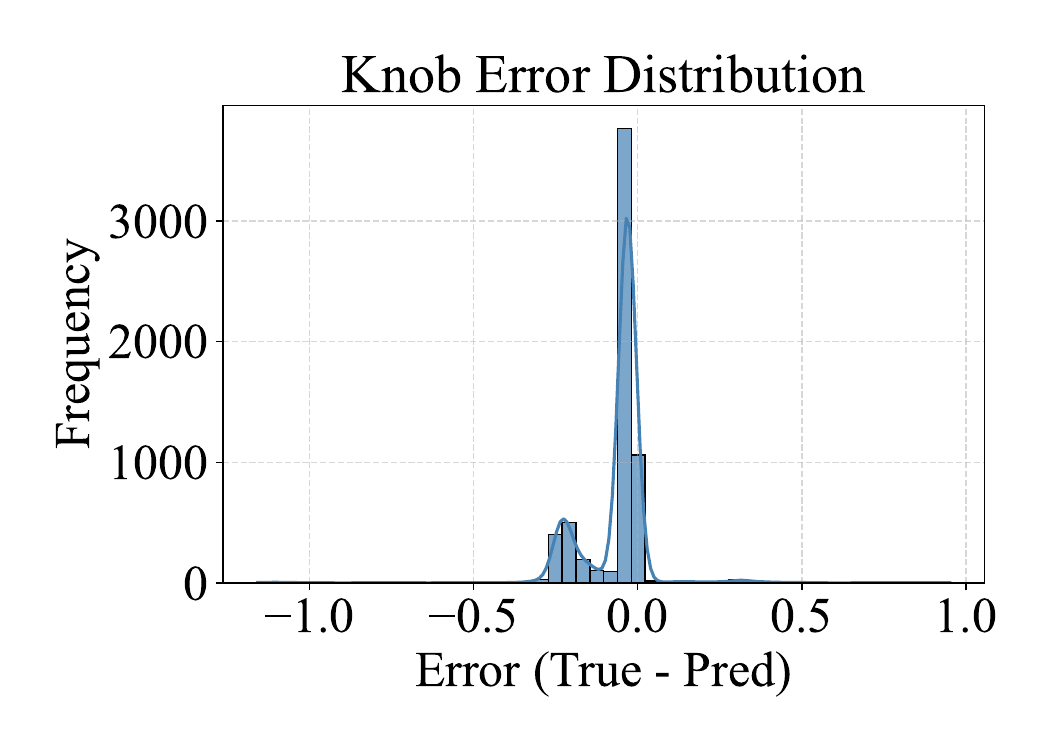}\label{fig:sc_hist_knob}}
\hspace{0.01\textwidth}
\subfloat[Rotation Error Distribution]{\includegraphics[width=0.30\textwidth]{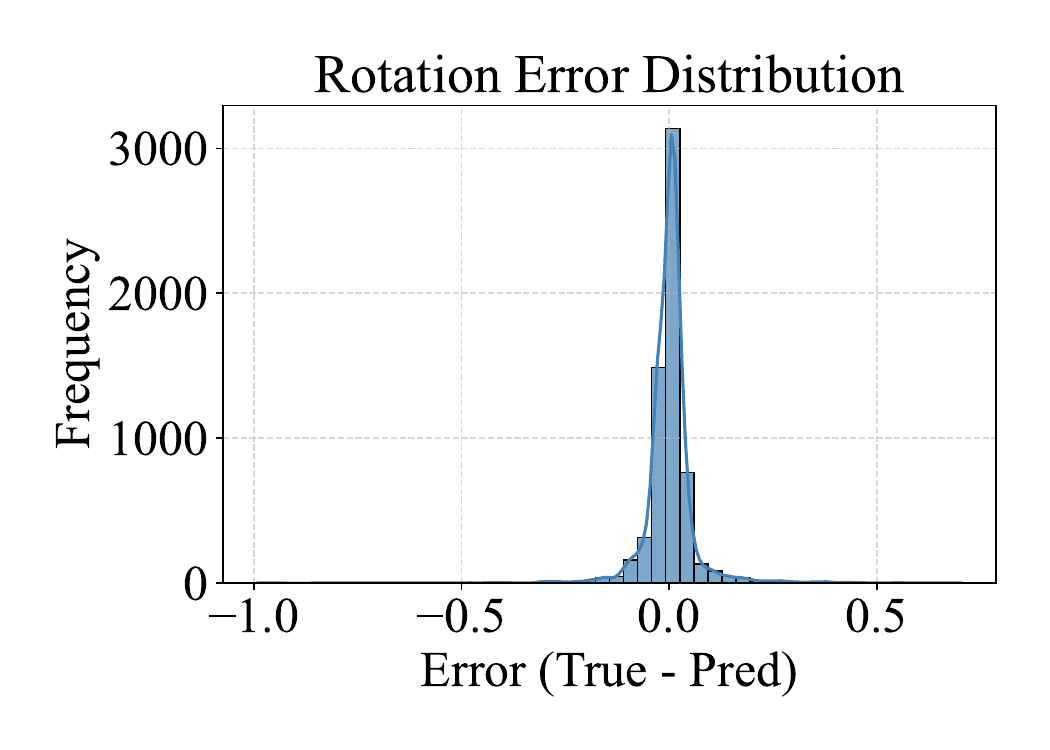}\label{fig:sc_hist_rot}}
\hspace{0.01\textwidth}
\subfloat[Translation Error Distribution]{\includegraphics[width=0.30\textwidth]{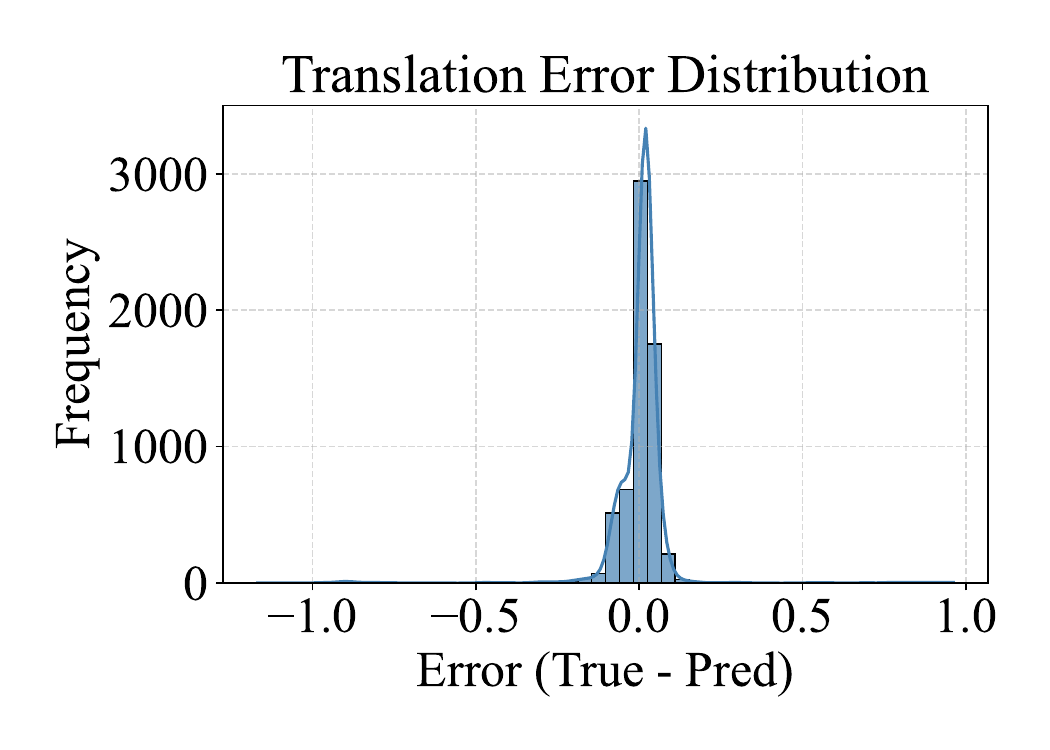}\label{fig:sc_hist_trans}}

\caption{True–predicted plots and error histograms for knob, rotation, and translation control using DINO-CVA. 
Episode-based results are shown in (a–f): (a) knob true–predicted, (b) rotation true–predicted, (c) translation true–predicted, (d) knob error distribution, (e) rotation error distribution, and (f) translation error distribution. 
Scenario-based results are shown in (g–l): (g) knob true–predicted, (h) rotation true–predicted, (i) translation true–predicted, (j) knob error distribution, (k) rotation error distribution, and (l) translation error distribution. 
The true–predicted plots illustrate alignment between predicted and ground-truth motor states, while the histograms depict the distribution of prediction errors across tasks.}
\label{fig:results}
\end{figure*}

\subsection{Results}
As mentioned, DINO-CVA was trained on two distinct training sets and validated on their corresponding test sets to evaluate its performance in steering the catheter by estimating joystick states toward partially unseen and entirely unseen destinations. Since the state vectors $s_{t+1}$ are generated in an autoregressive manner, the model outputs were evaluated using Mean Squared Error (MSE), Root Mean Squared Error (RMSE), Mean Absolute Error (MAE), and the coefficient of determination ($R^2$). To contextualize the results, a kinematics-only behavior cloning model (LSTM-based which was already used in \cite{farzad, farzad_2, fekri1, fekri2}) was implemented as a baseline, allowing us to assess whether the proposed multimodal framework can achieve comparable performance while also being visually aware.
Table \ref{tab:results} summarizes the performance of DINO-CVA across the episode-based and scenario-based dataset partitions. 
The results across both episode-based and scenario-based settings show consistently high accuracy for both models. In the episode-based evaluation, translation and rotation tasks yielded low error values and strong $R^2$ scores for both approaches, while knob control remained the most challenging but still achieved reliable predictions. A similar trend is observed in the scenario-based setting, where both models maintained high performance with only modest differences across metrics. These outcomes confirm that purely temporal modeling of kinematic sequences, as performed by the LSTM, provides a strong baseline for predicting short-term joystick states. However, the goal of our model is not to surpass the LSTM on raw error values, but to demonstrate that comparable performance can be achieved while projecting visual and kinematic information into a joint embedding space. This results in a vision-aware and kinematics-aware policy, in contrast to the LSTM baseline, which only models the sequence of joystick inputs without considering the spatial environment or vessel trajectory.

\par
In fact, the central distinction of our framework lies not in marginal metric improvements but in the representation it learns. The proposed model projects visual observations and kinematic sequences into a joint embedding space, where the two modalities are entangled to form a vision-aware and kinematics-aware policy. In contrast, the LSTM baseline relies solely on temporal continuity of joystick sequences and does not attend to the spatial environment or vascular geometry. As a result, while the LSTM can correctly predict actions whenever the kinematic sequence is consistent, it may still fail to ensure that the catheter follows the correct anatomical trajectory. Our model, by incorporating both visual and temporal cues, integrates the notion of trajectory into decision-making and learns to associate expert demonstrations with the vascular context in which they occur.
This difference is crucial in the context of autonomous navigation,  beyond reproducing joystick sequences, the multimodal fusion enables the policy to remain aware of where the catheter is within the vasculature, and where it should go. Thus, while the performance values appear close, the proposed approach takes a necessary step toward embedding trajectory awareness into the control process that purely kinematic models inherently lack.
\par
\par
Furthermore, the true-versus-predicted plots together with the error histograms provide a detailed view of model behavior across translation, rotation, and knob states in both dataset settings, illustrating not only the overall accuracy but also the distribution of prediction errors. These visual analyses further corroborate the quantitative findings, showing that the model produces reliable and stable predictions with bounded error distributions. In the following section, the results of the ablation study are presented to examine the contribution of the goal-conditioning head and the two-modality fusion strategy, thereby clarifying their impact on overall performance.

\begin{table}[ht]
\centering
\caption{Ablation study on the scenario-based dataset. The table reports the performance of the model and its variants when goal or modality components are removed.}
\label{tab:ablation}
\begin{tabular}{llcccc}
\toprule
Condition & Task & MSE & RMSE & MAE & $R^2$ \\
\midrule
\multirow{3}{*}{Baseline} 
  & Translation & 0.0116 & 0.1078 & 0.0369 & 0.9320 \\
  & Rotation    & 0.0050 & 0.0708 & 0.0354 & 0.8858 \\
  & Knob        & 0.0187  & 0.1367  & 0.0490  & 0.9116 \\
\midrule
\multirow{3}{*}{False goal} 
  & Translation & 0.1134 & 0.3368 & 0.2996 & 0.3364 \\
  & Rotation    & 0.0113 & 0.1061 & 0.0637 & 0.7437 \\
  & Knob        & 0.0616 & 0.2483 & 0.1333 & 0.6102 \\
\midrule
\multirow{3}{*}{No Goal} 
  & Translation & 0.0321 & 0.1792 & 0.1461 & 0.8120 \\
  & Rotation    & 0.0078 & 0.0881 & 0.0669 & 0.8234 \\
  & Knob        & 0.0456 & 0.2136 & 0.1395 & 0.7114 \\
\midrule
\multirow{3}{*}{No vision} 
  & Translation & 0.6098 & 0.7809 & 0.6943 & -2.5684 \\
  & Rotation    & 0.0447 & 0.2113 & 0.1322 & -0.0163 \\
  & Knob        & 0.1890 & 0.4347 & 0.2044 & -0.1951 \\
\midrule
\multirow{3}{*}{No states} 
  & Translation & 0.4273 & 0.6537 & 0.6108 & -1.5007 \\
  & Rotation    & 0.0580 & 0.2408 & 0.1599 & -0.3197 \\
  & Knob        & 0.1884 & 0.4341 & 0.1984 & -0.1916 \\
\bottomrule
\end{tabular}
\end{table}

\section{Ablation Study}
\label{ablation}
It is important to note that the proposed framework is grounded in imitation learning, specifically behavior cloning, which is inherently limited in terms of generalization and extrapolation beyond the training distribution. Nonetheless, the incorporation of goal conditioning is expected to mitigate this limitation.
To further investigate the contribution of goal conditioning and multimodal fusion, an ablation study was conducted on the scenario-based dataset, which represents the most challenging setting by requiring generalization to entirely unseen destinations (Table \ref{tab:ablation}). The first row corresponds to the baseline DINO-CVA performance reported previously in Table \ref{tab:results}. The impact of goal conditioning was first examined. When a false goal image was provided, performance degraded dramatically across all tasks, with $R^2$ values dropping to 0.336 for translation, 0.744 for rotation, and 0.610 for knob control. This highlights the central role of correct goal specification in guiding the model’s predictions, as misleading goal embeddings can cause the network to produce erroneous control outputs. When the goal head was entirely deactivated, performance also deteriorated, though less severely than in the false goal case (e.g., $R^2 = 0.812$ for translation and $R^2 = 0.711$ for knob), demonstrating that goal conditioning contributes positively to trajectory planning, particularly for fine-grained knob control.
\par
The final two conditions evaluated the role of modality fusion by removing either vision or state inputs. Eliminating the visual modality resulted in substantial performance collapse, especially for translation ($R^2 = -2.57$), underscoring that visual feedback is indispensable for spatial awareness and trajectory alignment. Similarly, removing the state inputs led to large performance drops across all tasks (e.g., $R^2 = -1.50$ for translation), indicating that motor state information is also essential for effective control. Overall, the ablation results confirm that both modalities are necessary and complementary: vision provides global spatial context, while joystick states encode dynamic control signals. Furthermore, the goal-conditioning head is shown to be critical for accurate navigation toward unseen destinations, with its absence or mis-specification leading to significant degradation in performance. These findings emphasize that the combination of multimodal fusion with goal conditioning is a key factor in enabling robust catheter steering in novel environments.
\section{Limitations}
\label{limitations}
This work is towards exploring the feasibility of multimodal, goal-conditioned architectures for autonomous catheter navigation, rather than presenting a definitive solution. Several limitations should therefore be acknowledged. First, the model currently employs fixed and learnable positional embeddings, which restricted our ability to evaluate performance under varying sequence lengths. Exploring alternative positional encoding strategies would be essential for assessing robustness to temporal variations. Second, the goal representation was incorporated through the CLS token of the goal image, rather than through a richer spatial embedding integrated into the gating mechanism. Although effective, this design choice leaves unexplored the potential benefits of spatially-aware goal conditioning. Third, the evaluation setup was limited by the available vascular phantom. The effect of visual awareness and goal conditioning should ideally be examined in a broader anatomical context with diverse branching structures, to better evaluate the ability of the model to correctly disambiguate trajectories when multiple vascular branches are present. Similarly, incorporating explicit destination coordinates into the goal conditioning, or marking the destination with visual cues (e.g., an index line or red dot in the goal image), could further enhance goal-driven navigation. Finally, the training and evaluation were performed on synthetic RGB data generated from the phantom setup, while clinical catheterization is guided by X-ray fluoroscopy. Extending the study to synthetic X-ray data would provide a more realistic representation of clinical conditions and help bridge the gap between laboratory evaluation and clinical applicability.
\section{Conclusions}
\label{conclude}
This work presented DINO-CVA, a multimodal goal-conditioned behavior cloning framework for autonomous catheter navigation. By jointly embedding visual and kinematic modalities, the model learns expert demonstrations in a trajectory and vision-aware manner. Training and evaluating on data obtained from a robotic experimental setup with a synthetic vascular phantom showed that the model achieves high accuracy across both episode and scenario-based datasets, performing on par with a kinematics-only LSTM while additionally grounding predictions in the anatomical context.
Ablation studies further demonstrated that the entanglement of goal conditioning and multimodal representations enhances model performance, particularly for fine-grained knob steering. These findings highlight the feasibility of multimodal, goal-conditioned architectures as a step towards reducing operator dependency and improving the reliability of catheter-based therapies.
Future work will focus on extending to synthetic X-ray data, incorporating richer spatial goal embeddings and destination markers, exploring more complex vascular phantoms, and incorporating simulation environments to complement the robotic platform for scalable training and evaluation.

\bibliographystyle{IEEEtran}

\end{document}